\documentclass[11pt]{article} %
\usepackage{times}
\usepackage{url}            %
\usepackage{booktabs}       %
\usepackage{amsfonts}       %
\usepackage{nicefrac}       %
\usepackage{microtype}      %
\usepackage{mkolar_definitions}
\usepackage{xspace}
\usepackage{multirow}
\usepackage{amsmath}
\usepackage{algorithm}
\usepackage{algorithmic}
\usepackage{color}
\usepackage{color, colortbl}
\usepackage{enumitem}
\usepackage{comment}
\usepackage{bm}
\usepackage[left=2cm,top=2cm,right=2cm]{geometry}

\usepackage[scaled=.9]{helvet}

\usepackage{url}
\usepackage{authblk}
\usepackage{csquotes}

\usepackage[pdftex]{graphicx}

\usepackage[dvipsnames]{xcolor}
\usepackage[colorlinks=true, linkcolor=blue, citecolor=blue, urlcolor=WildStrawberry]{hyperref}

\usepackage{natbib}
\usepackage{booktabs}       %
\usepackage{amsfonts}       %
\usepackage{nicefrac}       %
\usepackage{microtype}      %
\usepackage{mkolar_definitions}

\usepackage{enumitem}      

\newcommand\numberthis{\addtocounter{equation}{1}\tag{\theequation}}
\allowdisplaybreaks

\newcommand{\ls}{\ell}

\newcommand{\ldef}{\vcentcolon=}
\newcommand{\rdef}{=\vcentcolon}

\newcommand{\wt}[1]{\widetilde{#1}}
\newcommand{\wh}[1]{\widehat{#1}}

\def\ddefloop#1{\ifx\ddefloop#1\else\ddef{#1}\expandafter\ddefloop\fi}
\def\ddef#1{\expandafter\def\csname bb#1\endcsname{\ensuremath{\mathbb{#1}}}}
\ddefloop ABCDEFGHIJKLMNOPQRSTUVWXYZ\ddefloop
\def\ddefloop#1{\ifx\ddefloop#1\else\ddef{#1}\expandafter\ddefloop\fi}
\def\ddef#1{\expandafter\def\csname b#1\endcsname{\ensuremath{\mathbf{#1}}}}
\ddefloop ABCDEFGHIJKLMNOPQRSTUVWXYZ\ddefloop
\def\ddef#1{\expandafter\def\csname c#1\endcsname{\ensuremath{\mathcal{#1}}}}
\ddefloop ABCDEFGHIJKLMNOPQRSTUVWXYZ\ddefloop
\def\ddef#1{\expandafter\def\csname h#1\endcsname{\ensuremath{\widehat{#1}}}}
\ddefloop ABCDEFGHIJKLMNOPQRSTUVWXYZabcdefghijklmnopqrsuvwxyz\ddefloop    
\def\ddef#1{\expandafter\def\csname hc#1\endcsname{\ensuremath{\widehat{\mathcal{#1}}}}}
\ddefloop ABCDEFGHIJKLMNOPQRSTUVWXYZ\ddefloop
\def\ddef#1{\expandafter\def\csname t#1\endcsname{\ensuremath{\widetilde{#1}}}}
\ddefloop ABCDEFGHIJKLMNOPQRSTUVWXYZ\ddefloop
\def\ddef#1{\expandafter\def\csname tc#1\endcsname{\ensuremath{\widetilde{\mathcal{#1}}}}}
\ddefloop ABCDEFGHIJKLMNOPQRSTUVWXYZ\ddefloop

\usepackage{tikz}

\newcommand{\proman}[1]{\prn*{\romannumeral #1}}

\newcommand{\overleq}[1]{\overset{ #1}{\leq{}}}

\newcommand{\overeq}[1]{\overset{#1}{=}}

\newtheorem*{theorem*}{Theorem}

\newcommand{\dynamicsopt}{\cT}
\newcommand{\transitionopt}{P}
\newcommand{\Vmax}{V_{\mathrm{max}}}
\newcommand{\algname}{Hy-Q} 

\newcommand{\moff}{m_{\mathrm{off}}}
\newcommand{\mon}{m_{\mathrm{on}}}

\newcommand{\Ball}{\bbB}

\newcommand{\distrib}{\rho}

\usepackage{xspace} 
\usepackage{multirow}
\usepackage{amsmath} 
\usepackage{algorithm}
\usepackage{algorithmic}
\usepackage{color}
\usepackage{color, colortbl}
\usepackage{enumitem}
\usepackage{comment}
\usepackage{bm}
\usepackage{caption}
\usepackage{subcaption}

\usepackage{url}
\usepackage{csquotes}

\usepackage{centernot}

\newcount\Comments  %
\definecolor{darkgreen}{rgb}{0,0.5,0}
\definecolor{darkred}{rgb}{0.7,0,0}
\definecolor{teal}{rgb}{0.3,0.8,0.8}
\definecolor{orange}{rgb}{1.0,0.5,0.0}
\definecolor{purple}{rgb}{0.8,0.0,0.8}
\newcommand{\kibitz}[2]{\ifnum\Comments=1{\textcolor{#1}{\textsf{\footnotesize #2}}}\fi}

\usepackage{color-edits}

\definecolor{Gray}{gray}{0.9}

\newcommand{\algcomment}[1]{\textcolor{blue!70!black}{\footnotesize{\texttt{\textbf{//
          #1}}}}}

\newcommand{\briee}{\textsc{Briee}\xspace}
\newcommand{\rnd}{\textsc{Rnd}\xspace}
\newcommand{\cql}{\textsc{Cql}\xspace}

\newcommand{\dqfd}{\textsc{Dqfd}\xspace}

\author{Yuda Song\thanks{Carnegie Mellon University. Email: \texttt{yudas@cs.cmu.edu}}\; Yifei Zhou\footnote{Cornell University. Email: \texttt{yz639@cornell.edu}. The first two authors have equal contributions to the paper.} \; Ayush Sekhari\footnote{MIT. Email: \texttt{sekhari@mit.edu}} \; J. Andrew Bagnell\footnote{Carnegie Mellon University. Email: \texttt{dbagnell@aurora.tech}} \; Akshay Krishnamurthy\footnote{Microsoft Research. Email: \texttt{akshaykr@microsoft.com}} \; Wen Sun\footnote{Cornell University. Email: \texttt{ws455@cornell.edu}}}

\title{Hybrid RL: Using Both Offline and Online Data Can Make RL Efficient} 

\usepackage{mathrsfs}

\begin{document} 

\maketitle
\begin{abstract}
We consider a hybrid reinforcement learning setting (Hybrid RL), in which an agent has access to an offline dataset and the ability to collect experience via real-world online interaction. The framework mitigates the challenges that arise in both pure offline and online RL settings, allowing for the design of simple and highly effective algorithms, in both theory and practice. We demonstrate these advantages by adapting the classical Q learning/iteration algorithm to the hybrid setting, which we call Hybrid Q-Learning or Hy-Q. In our theoretical results, we prove that the algorithm is both computationally and statistically efficient whenever the offline dataset supports a high-quality policy and the environment has bounded bilinear rank. Notably, we require no assumptions on the coverage provided by the initial distribution, in contrast with guarantees for policy gradient/iteration methods. In our experimental results, we show that Hy-Q with neural network function approximation outperforms state-of-the-art online, offline, and hybrid RL baselines on challenging benchmarks, including Montezuma's Revenge. 
\end{abstract}

\section{Introduction}

Learning by interacting with an environment, in the standard online reinforcement learning (RL) protocol, has led to impressive results across a number of domains. State-of-the-art RL algorithms are quite general, employing function approximation to scale to complex environments with minimal domain expertise and inductive bias. However, online RL agents are also notoriously sample inefficient, often requiring billions of environment interactions to achieve suitable performance. This issue is particularly salient when the environment requires sophisticated exploration and a high quality reset distribution is unavailable to help overcome the exploration challenge. 
As a consequence, the practical success of online RL and related policy gradient/improvement methods has been largely restricted to settings where a high quality simulator is available. 

To overcome the issue of sample inefficiency, attention has turned to the offline RL setting~\citep{levine2020offline}, where, rather than interacting with the environment, the agent trains on a large dataset of experience collected in some other manner (e.g., by a system running in production or an expert). While these methods still require a large dataset, they mitigate the sample complexity concerns of online RL, since the dataset can be collected without compromising system performance. However, offline RL methods can suffer from \emph{distribution shift}, where the state distribution induced by the learned policy differs significantly from the offline distribution~\citep{wang2021instabilities}. Existing provable approaches for addressing distribution shift are computationally intractable, while empirical approaches rely on heuristics that can be sensitive to the domain and offline dataset (as we will see).

\begin{figure}[h]
  \begin{center}
    \includegraphics[width=0.45\textwidth]{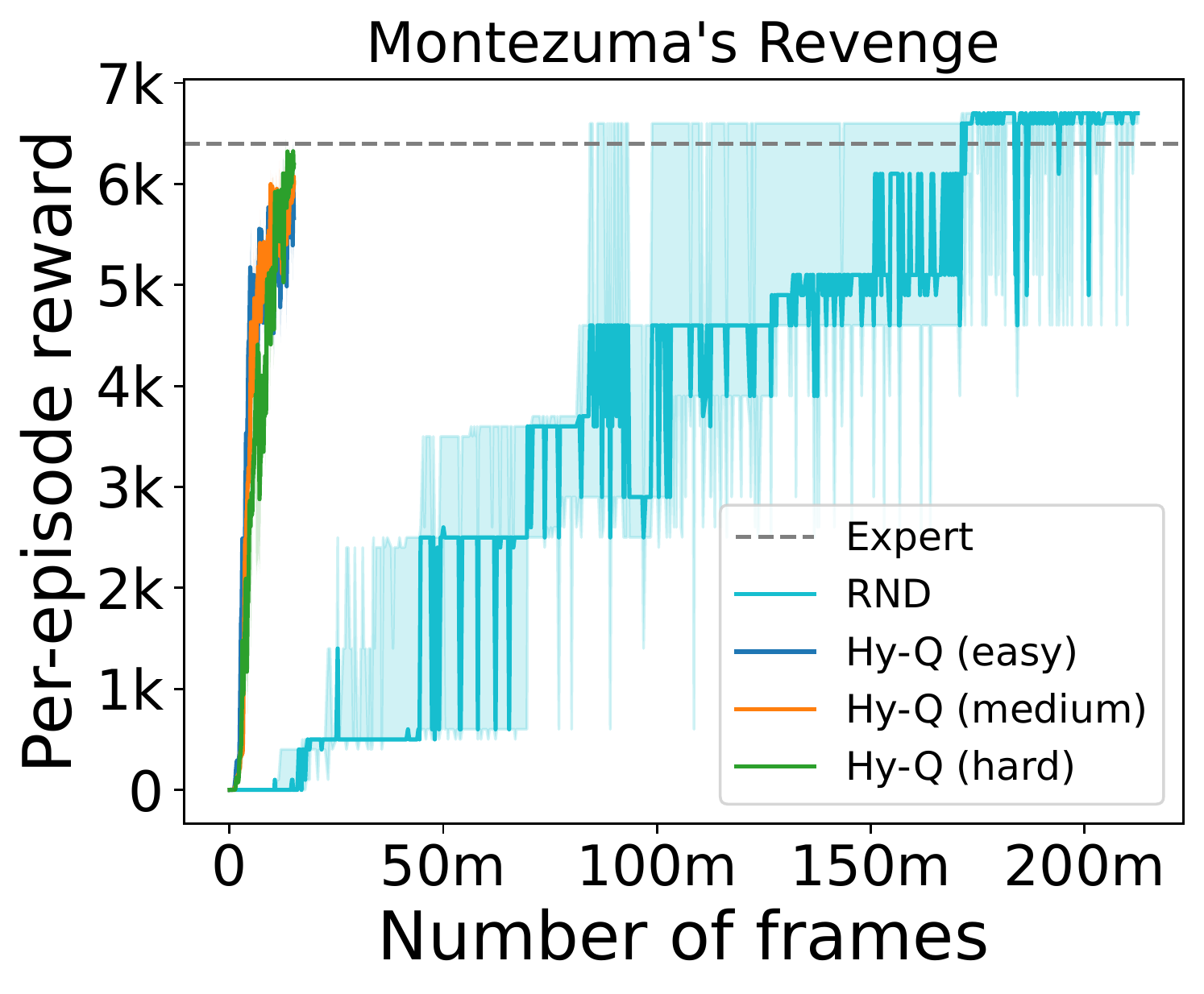}
  \end{center}
  \caption{Performance of our approach \algname{}  on Montezuma's Revenge. We consider three types of offline datasets: easy, medium, and hard. The easy one contains offline data from a high quality policy (denoted by \emph{Expert} in the figure), the medium one consists of 20\% data from a random policy and 80\% from the expert, and the hard one consists of half random data and half expert data. All offline datasets have 100k tuples of state, action, reward, and next state (no trajectory-wise information is available to the learners). With only 0.1m offline samples, our approach \algname{} learns nearly 10x faster than \emph{Random Network Distillation (RND)} -- an online deep RL baseline designed for  Montezuma's revenge (see \pref{fig:mrevenge_visual} for a visual comparison using screenshots of the game). We also note that \emph{Conservative Q-Learning (CQL)}---an offline RL baseline, completely fails in all three offline datasets, indicating the hardness of  offline policy learning from such small size offline datasets. Imitation Learning baseline \emph{Behavior Cloning (BC)} achieves reasonable performance in easy and medium datasets, but fails completely on the hard dataset. See \pref{sec:m_revenge} for more detailed comparison to CQL, BC, and other baselines.}
  \label{fig:rnd}
\end{figure}

In this paper, we focus on a hybrid reinforcement learning setting, which we call Hybrid RL, that draws on the favorable properties of both offline and online settings. In Hybrid RL, the agent has both an offline dataset and the ability to interact with the environment, as in the traditional online RL setting. The offline dataset helps address the exploration challenge, allowing us to greatly reduce the number of interactions required. Simultaneously, we can identify and correct distribution shift issues via online interaction. Variants of the setting have been studied in a number of empirical works~\citep{rajeswaran2017learning,hester2018deep,nair2018overcoming,nair2020accelerating, vecerik2017leveraging} which mainly focus on using  expert demonstrations as offline data.

Hybrid RL is closely related to the \emph{reset setting}, where the agent can interact with the environment starting from a ``nice" distribution. A number of simple and effective algorithms, including CPI \citep{kakade2002approximately}, PSDP \citep{bagnell2003policy}, and policy gradient methods \citep{kakade2001natural,agarwal2020optimality}
-- methods which have inspired recent powerful heuristic RL methods such as TRPO \citep{schulman2015trust} and PPO \citep{schulman2017proximal}---
are provably efficient in the reset setting. Yet, leveraging a reset distribution is a strong requirement (often tantamount to having access to a detailed simulation) and unlikely to be available in real world applications. Hybrid RL differs from the reset setting in that (a) we have an offline dataset, but (b) our online interactions must come from traces that start at the initial state distribution of the environment, and the initial state distribution is not assumed to have any nice properties.
Both features (offline data and a nice reset distribution) facilitate algorithm design by de-emphasizing the exploration challenge. However, Hybrid RL is more practical since an offline dataset is much easier to access, while a nice reset distribution or even generative model is generally not guaranteed in practice. \looseness=-1 

We showcase the Hybrid RL setting with a new algorithm, Hybrid Q learning or Hy-Q (pronounced: Haiku). The algorithm is a simple adaptation of the classical fitted Q-iteration algorithm (FQI) and it accommodates value-based function approximation in a relatively general setup. 
\footnote{We use Q-learning and Q-iteration interchangeably, although they are not strictly speaking the same algorithm. Our theoretical results analyze Q-iteration, but we use an algorithm with an online/mini-batch flavor that is closer to Q-learning for our experiments.} For our theoretical results, we prove that Hy-Q is both statistically and computationally efficient assuming that: (1) the offline  distribution covers some high quality policy, (2) the MDP has low bilinear rank, (3) the function approximator is Bellman complete, and (4) we have a least squares regression oracle. The first three assumptions are standard statistical assumptions in the RL literature while the fourth is a widely used computational abstraction for supervised learning. No computationally efficient algorithms are known under these assumptions in pure offline or pure online settings, which highlights the advantages of the hybrid setting.

We also implement \algname{} and evaluate it on two challenging RL benchmarks: a rich observation combination lock \citep{misra2020kinematic} and Montezuma's Revenge from the Arcade Learning Environment \citep{bellemare2013arcade}. Starting with an offline dataset that contains some transitions from a high quality policy, our approach outperforms: an online RL baseline with theoretical guarantees, an online deep RL baseline tuned for Montezuma's Revenge, a pure offline RL baseline, an imitation learning baseline, and an existing hybrid method. Compared to the online methods, Hy-Q requires only a small fraction of the online experience, demonstrating its sample efficiency (e.g., \pref{fig:rnd}). Compared to the offline and hybrid methods, Hy-Q performs most favorably when the offline dataset also contains many interactions from low quality policies, demonstrating its robustness. These results reveal the significant benefits that can be realized by combining offline and online data.

\section{Related Works}
We discuss related works from four categories: pure online RL, online RL with access to a reset distribution, offline RL, and prior work in hybrid settings. We note that pure online RL refers to the setting where one can only reset the system to initial state distribution of the environment, which is not assumed to provide any form of coverage. 

\paragraph{Pure online RL}
Beyond tabular settings, many existing statistically efficient RL algorithms are not computationally tractable, due to the difficulty of implementing optimism. 
This is true in the linear MDP~\citep{jin2020provably} with large action spaces, the linear Bellman complete model~\citep{zanette2020learning,agarwal2019reinforcement}, and in the general function approximation setting~\citep{jiang2017contextual,sun2019model,du2021bilinear,jin2021bellman}.
These computational challenges have inspired results on intractability of aspects of online RL~\citep{dann2018oracle,kane2022computational}. On the other hand, many simple exploration based algorithms like \(\epsilon\)-greedy are  computationally efficient, but they may not always work well in practice. Recent theoretical works  \citep{dann2022guarantees, liu2018simple} have explored the additional structural assumptions on the underlying dynamics and value function class under which \(\epsilon\)-greedy succeeds, but they still do not capture all the relevant practical problems. 

There are several online RL algorithms that aim to tackle the computational issue via stronger structural assumptions and supervised learning-style computational oracles~\citep{misra2020kinematic, sekhari2021agnostic, zhang2022efficient,agarwal2020flambe,uehara2021representation,modi2021model,zhang2022making,qiu2022contrastive}.
Compared to these oracle-based methods, our approach operates in the more general ``bilinear rank" setting and relies on a standard supervised learning primitive: least squares regression. Notably, our oracle admits efficient implementation with linear function approximation, so we obtain an end-to-end computational guarantee; this is not true for prior oracle-based methods. 

There are many deep RL methods for the online setting (e.g.,~\citet{schulman2015trust,schulman2017proximal,lillicrap2015continuous,haarnoja2018soft,schrittwieser2020mastering}). Apart from a few exceptions (e.g., \citet{burda2018exploration,badia2020agent57,guo2022byol}), most rely on random exploration (e.g. \(\epsilon\)-greedy) and are not capable of strategic exploration. In fact, guarantees for \(\epsilon\)-greedy like algorithms only exist under additional structural assumptions on the underlying problem. 

In our experiments, we test our approach on Montezuma's Revenge, and we pick \textsc{Rnd} \citep{burda2018exploration} as a deep RL exploration baseline due to its simplicity and effectiveness on the game of Montezuma's Revenge. 

\paragraph{Online RL with reset distributions}
When an exploratory reset distribution is available, a number of statistically and computationally efficient algorithms are known. The classic algorithms are \textsc{Cpi} \citep{kakade2002approximately}, \textsc{Psdp} \citep{bagnell2003policy}, Natural Policy Gradient \citep{kakade2001natural,agarwal2020optimality}, and \textsc{PolyTex} \citep{abbasi2019politex}. 
\cite{uchendu2022jump} recently demonstrated that algorithms like \textsc{Psdp} work well when equipped with modern neural network function approximators. However, these algorithms (and their analyses) heavily rely on the reset distribution to mitigate the exploration challenge, but such a reset distribution is typically unavailable in practice, unless one also has a simulator and access to its internal states. 
In contrast, we assume the offline data covers some high quality policy (no need to be globally exploratory), which helps with exploration, but we do not require an exploratory reset distribution. 
This makes the hybrid setting much more practically appealing.

\paragraph{Offline RL}
Offline RL methods learn policies solely from a given offline dataset, with no interaction whatsoever. 
When the dataset has global coverage,  algorithms such as FQI~\citep{munos2008finite,chen2019information} or certainty-equivalence model learning~\citep{ross2012agnostic}, can find near-optimal policies in an oracle-efficient manner, via least squares or model-fitting oracles. However, with only partial coverage, existing methods either (a) are not computationally efficient due to the difficulty of implementing pessimism both in linear settings with large action spaces~\citep{jin2021pessimism,zhang2022corruption,chang2021mitigating} and general function approximation settings~\citep{uehara2021pessimistic,xie2021bellman,jiang2020minimax,chen2022offline,zhan2022offline}, or (b) require strong representation conditions such as policy-based Bellman completeness~\citep{xie2021bellman,zanette2021provable}. In contrast, in the hybrid setting, we obtain an efficient algorithm under the more natural condition of completeness w.r.t., the Bellman optimality operator only. 

Among the many empirical offline RL methods (e.g., \cite{kumar2020conservative,yu2021combo,kostrikov2021offline,fujimoto2021minimalist}), we use \textsc{Cql}~\citep{kumar2020conservative} as a baseline in our experiments, since it has been shown to work in image-based settings such as Atari games. 

\paragraph{Online RL with offline datasets} 
\cite{ross2012agnostic} developed a model-based algorithm for a similar hybrid setting. In comparison, our approach is model-free and consequently may be more suitable for high-dimensional state spaces (e.g., raw-pixel images). \cite{xie2021policy} studied hybrid RL and show that offline data does not yield statistical improvements in tabular MDPs. 
Our work instead focuses on the function approximation setting and demonstrates computational benefits of hybrid RL. 

On the empirical side, several works consider combining offline expert demonstrations with online interaction \citep{rajeswaran2017learning,hester2018deep,nair2018overcoming,nair2020accelerating, vecerik2017leveraging}. A common challenge in offline RL is the robustness against low-quality offline dataset. Previous works mostly focus on expert demonstrations and have no rigorous guarantees for such robustness. In fact, \citet{nair2020accelerating} showed that such degradation in performance indeed happens in practice with low-quality offline data. In our experiments, we observe that DQfD \citep{hester2018deep} also has a similar degradation. On the other hand, our algorithm is robust to the quality of the offline data. Note that the core idea of our algorithm is similar to that of \citet{vecerik2017leveraging}, who adapt DDPG to the setting of combining RL with expert demonstrations for continuous control. Although \citet{vecerik2017leveraging} does not provide any theoretical results, it may be possible to combine our theoretical insights with existing analyses for policy gradient methods to establish some guarantees of the algorithm from \citet{vecerik2017leveraging} for the hybrid RL setting. We also include a detailed comparison with previous empirical work in \pref{app:comp}. \looseness=-1

\section{Preliminaries}
 We consider finite horizon Markov Decision Process $M(\Scal, \Acal, H, R, \transitionopt, d_0)$, where $\Scal$ is the state space, $\Acal$ is the action space, $H$ denotes the horizon, stochastic rewards $R(s,a) \in \Delta([0,1])$ and $\transitionopt(s,a) \in \Delta(\Scal)$ are the reward and transition distributions at $(s,a)$, and $d_0 \in \Delta(\Scal)$ is the initial distribution. We assume the agent can only reset from $d_0$ (at the beginning of each episode). Since the optimal policy is non-stationary in this setting, we define a policy $\pi := \{\pi_0, \dots, \pi_{H-1}\}$ where $\pi_h:\Scal\mapsto \Delta(\Acal)$. Given $\pi$, $d^{\pi}_h\in \Delta(\Scal\times\Acal)$ denotes the state-action occupancy induced by $\pi$ at step $h$. \looseness=-1
Given $\pi$, we define the state and state-action value functions in the usual manner: $V^\pi_h(s) = \EE[ \sum_{\tau = h}^{H-1} r_\tau | \pi, s_h = s ]$ and $Q^\pi_h(s,a) = \EE[ \sum_{\tau = h}^{H-1} r_\tau | \pi, s_h = s, a_h=a]$.
$Q^\star$ and $V^\star$ denote the optimal value functions. We denote $V^\pi = \EE_{s_0\sim d_0} V^{\pi}_0(s_0)$ as the expected total reward of $\pi$. 
We define the Bellman operator $\dynamicsopt$ such that for any $f: \Scal\times\Acal\mapsto \mathbb{R}$, 
$$\dynamicsopt f(s,a) = \EE[R(s,a)] + \EE_{s'\sim \transitionopt(s,a)} \max_{a'} f(s',a') \qquad \forall s,a, $$
We assume that for each $h$ we have an offline dataset of $\moff$ samples $(s,a,r,s')$ drawn iid via $(s,a) \sim \nu_h, r \in R(s,a), s' \sim P(s,a)$. Here $
\nu = \{\nu_0, \dots, \nu_{H-1}\}$ denote the corresponding offline data distributions.  
For a dataset $\Dcal$, we use $\hat{\EE}_\Dcal[\cdot]$ to denote a sample average over this dataset.
For our theoretical results, we will assume that $\nu$ covers some high-quality policy.  Note that covering a high quality policy does not mean that $\nu$ itself is a distribution of some high quality policy. For instance, $\nu$ could be a mixture distribution of some high quality policy and a few low-quality policies, in which case, treating $\nu$ as expert demonstration will fail completely (as we will show in our experiments).
We consider the value-based function approximation setting, where we are given a function class $\Fcal = \Fcal_0 \times \dots \times \cF_{H-1}$ with $\Fcal_h \subset \Scal\times\Acal\mapsto [0, \Vmax]$ that we use to approximate the value functions for the underlying MDP. Here $\Vmax$ denotes the maximum total reward of a trajectory. For ease of notation, we define $f = \{f_0,\ldots,f_{H-1}\}$ and define $\pi^f$ to be the greedy policy w.r.t., $f$, which chooses actions as $\pi_h^f(s) = \argmax_a f_h(s,a)$.

\section{Hybrid Q-Learning} 
\label{sec:hyq}
\begin{algorithm}[t] 
\caption{Hybrid Q-learning using both offline and online data (Hy-Q)}
\begin{algorithmic}[1] 
\REQUIRE Value class: $\Fcal$, \#iterations: $T$, offline dataset \(\cD^{\nu}_h\) of size \(m_\mathrm{off}\) $=T$ for \(h \in [H-1]\). 
\\
\STATE Initialize \(f_h^1(s, a) = 0\). 
\FOR{$t = 1, \dots, T$}  
\STATE Let $\pi^t$ be the greedy policy w.r.t. \(f^t\) i.e., $\pi_h^t(s) = \argmax_a f^t_h(s,a).$ 
\STATE For each $h$, collect $m_{\mathrm{on}}$ $=1$ online tuples $\cD^t_h \sim d_h^{\pi^t}$.   \algcomment{Online collection} \\ 
 \vspace{1mm} 
 \algcomment{FQI using both online and offline data}
 \vspace{1mm}
 \label{line:online_sample}
\STATE Set $f_H^{t+1}(s,a) = 0$.  \\ 
\FOR{$h = H-1, \dots, 0$}  \label{line:fqi_iteration}
\STATE Estimate \(f_h^{t+1}\) using least squares regression on the aggregated data \(\Dcal^t_h = \Dcal^\nu_{h} + \sum_{\tau = 1}^{t} \Dcal^\tau_{h}\): 
 \begin{align*}
\hspace{-10mm} f_{h}^{t+1} \leftarrow \argmin_{f\in \Fcal_h}   \crl*{ 
 \wh \EE_{\Dcal^t_h}( f(s,a) -   r -  \max_{a'} f^{t+1}_{h+1}(s', a') )^2} \numberthis \label{eq:alg_regression} 
\end{align*}
\vspace{-0.5cm}
\ENDFOR
\ENDFOR
\end{algorithmic}\label{alg:fqi} 
\end{algorithm} 
In this section, we present our algorithm \emph{Hybrid Q Learning} -- \algname{} in \pref{alg:fqi}.  \algname{} takes an offline dataset $\{\Dcal^\nu\}_{h=1}^H$ that contains $(s,a,r, s')$ tuples and a Q function class $\Fcal_h \subset \Scal\times\Acal\mapsto [0, H], h=1,\dots, H,$ as inputs, and outputs a policy that optimizes the given reward function. The algorithm is conceptually simple: it
iteratively executes the FQI procedure
(\pref{line:fqi_iteration}) using the offline dataset \emph{and} on-policy samples generated by the learned policies.

Specifically, at iteration $t$ and timestep $h$, we have an estimate $f_h^t$ of the $Q_h^\star$ function and we set $\pi_h^t$ to be the greedy policy for $f_h^t$. We execute $\pi_h^t$ to collect a dataset $\Dcal_h^t$ of online samples in~\pref{line:online_sample}. More formally, we sample $s_h \sim d^{\pi^t}_h, a_h \sim \pi_h^t(\cdot | s_h), s_{h+1}\sim P(\cdot | s_h,a_h)$ and add the tuple $(s_h,a_h, r_h, s_{h+1})$ to $\Dcal_h^t$.  Then we run FQI, the dynamic programming style algorithm on both the offline dataset $\Dcal^\nu_h$ and all previously collected online samples $\{\Dcal_h^\tau\}_{\tau=1}^t$. The FQI update works backward from time step $H$ to $0$ and computes $f_h^{t+1}$ via least squares regression with input $(s,a)$ and regression target $r+ \max_{a'} f_{h+1}^{t+1}(s',a')$.

Let us make several remarks, here we drop timestep $h$ for generality. Intuitively, the FQI updates in Hy-Q try to ensure that the estimate $f^t$ has small Bellman error under both the offline distribution $\nu$ and the online distributions $d^{\pi^t}$. The standard offline version of FQI ensures the former, but this alone is insufficient when the offline dataset has poor coverage. Indeed FQI may have poor performance in such cases~\citep[see examples in][]{zhan2022offline,chen2022offline}. The key insight in Hy-Q is to use online interaction to ensure that we also have small Bellman error on $d^{\pi^t}$. As we will see, the moment we find an $f^t$ that has small Bellman error on the offline distribution $\nu$ and \emph{its own greedy policy's distribution $d^{\pi^t}$}, FQI guarantees that $\pi^t$ will be at least as good as \emph{any} policy covered by $\nu$. This observation results in an explore-or-terminate phenomenon: either $f^t$ has small Bellman error on its distribution and we are done, or $d^{\pi^t}$ must be significantly different from distributions we have seen previously and we make progress. Crucially, no explicit exploration is required for this argument, which is precisely how we avoid the computational difficulties with implementing optimism. 

Another important point pertains to \emph{catastrophic forgetting}. 
We will see that the size of the offline dataset $\moff$ should be comparable to the total amount of online data $\{\Dcal_h^\tau\}_{\tau=1}^T$, so that the two terms in Eq.~\ref{eq:alg_regression} have similar weight and we ensure  low Bellman error on $\nu$ throughout the learning process.  In practice, we implement this by having all model updates use a fixed (significant) number  
of offline samples even as we collect more online data, so that we do not ``forget" the distribution $\nu$. This is quite different from warm-starting with $\Dcal^\nu$ and then switching to online RL, which may result in catastrophic forgetting due to a vanishing proportion of offline samples being used for model training as we collect more online samples. We note that this balancing scheme is analogous to and inspired by the one used by~\citet{ross2012agnostic} in the context of model-based RL with a reset distribution.  Previously, similar techniques have also been explored for various applications (for example, see Appendix F.3 of \citet{kalashnikov2018qt}). As in ~\citet{ross2012agnostic}, a key practical insight from our analysis is that the offline data should be used throughout training to avoid catastrophic forgetting.

\section{Theoretical Analysis: Low Bilinear Rank Models} \label{sec:low_bilinear_rank} 

In this section we present the main theoretical guarantees for Hy-Q. We start by stating the main assumptions and definitions for the function approximator, the offline data distribution, and the MDP. We state the key definitions and then provide some discussion.

\begin{assum}[Realizability and Bellman completeness]
\label{ass:Bellman_completeness}
For any $h$, we have $Q_h^\star \in \Fcal_h$. Additionally, for any $f_{h+1} \in \Fcal_{h+1}$, we have $\Tcal f_{h+1} \in \Fcal_h$. 
\end{assum}

\begin{definition}[Bellman error transfer coefficient]
For any policy $\pi$, define the transfer coefficient as
\begin{align*}
C_{\pi} := \max\crl*{0, ~ \max_{f\in \Fcal} \frac{\sum_{h=0}^{H-1} \EE_{s,a\sim d_h^{\pi} }  \left[ \dynamicsopt f_{h+1}(s,a) - f_h(s,a) \right] }{ \sqrt{ \sum_{h=0}^{H-1} \EE_{s,a\sim \nu_h }  \left(\dynamicsopt f_{h+1}(s,a) -  f_h(s,a)  \right)^2 }}}. \numberthis  \label{eq:concentrability_definition}
\end{align*}
\end{definition}

The transfer coefficient definition above is somewhat non-standard, but is actually weaker than related notions used in prior offline RL results. First, the average Bellman error appearing in the numerator is weaker than the squared Bellman error notion of~\citep{xie2021bellman}; a simple calculation shows that $C_\pi^2$ is upper bounded by their coefficient. Second, by using Bellman errors, both of these are bounded by notions involving density ratios~\citep{kakade2002approximately,munos2008finite,chen2019information}. Third, when function $f\in \Fcal$ is linear in some known feature which is the case for models such as linear MDPs, the above transfer coefficient can be refined to relative condition number defined using the features.  Finally, many works, particularly those that do not employ pessimism~\citep{munos2008finite,chen2019information}, require ``all-policy" analogs, which places a much stronger requirement on the offline data distribution $\nu$.  In contrast, we will only ask that $C_\pi$ is small for \emph{some} high-quality policy that we hope to compete with. 
In \pref{app:transfer_coefficient_bound}, we showcase that our transfer coefficient is weaker than any related notions used in prior works under various settings such as 
tabular MDPs, linear MDPs, low-rank MDPs, and MDPs with general value function approximation. 

\begin{definition}[{Bilinear model \citep{du2021bilinear}}] 
\label{def:bilinear_model}
We say that the MDP together with the function class $\Fcal$ is a bilinear model of rank \(d\) if for any \(h \in [H-1]\), there exist two (unknown) mappings $X_h,W_h:\Fcal\mapsto \mathbb{R}^d$ with $\max_{f}\| X_h(f)\|_2 \leq B_X$ and $\max_{f} \|W_h(f)\|_2 \leq B_W$ such that:
\begin{align*} 
\forall f, g\in \Fcal: \; \abs*{\EE_{s, a \sim d_h^{\pi^f}}\sbr{  g_h(s,a) - \dynamicsopt g_{h+1}(s,a) } } = \left\lvert \tri*{ X_h(f), W_h(g) }  \right\rvert.
\end{align*} 
\end{definition}

All concepts defined above are frequently used in the statistical analysis of RL methods with function approximation. Realizability is the most basic function approximation assumption, but is known to be insufficient for offline RL~\citep{foster2021offline} unless other strong assumptions hold~\citep{xie2021batch,zhan2022offline,chen2022offline}. Completeness is the most standard strengthening of realizability that is used routinely in both online~\citep{jin2021bellman} and offline RL~\citep{munos2008finite,chen2019information} and is known to hold in several settings including the linear MDP and the linear quadratic regulator. These assumptions ensure that the dynamic programming updates of FQI are stable in the presence of function approximation. 

Lastly, the bilinear model was developed in a series of works~\citep{jiang2017contextual,jin2021bellman,du2021bilinear} on sample efficient online RL.\footnote{\citet{jin2021bellman} consider the Bellman Eluder dimension, which is related but distinct from the Bilinear model. However, our proofs can be easily translated to this setting; see~\pref{app:Bellman_eluder} for more details.} The setting is known to capture a wide class of models including linear MDPs, linear Bellman complete models, low-rank MDPs, Linear Quadratic Regulators, reactive POMDPs, and more. As a technical note, the main paper focuses on the ``Q-type" version of the bilinear model, but the algorithm and proofs easily extend to the ``V-type" version. See~\pref{app:V_type_appendix} for details.

\begin{theorem}[Cumulative suboptimality]
\label{thm:main_regret} 
Fix $\delta\in (0,1)$, \(\moff = T\) and \(\mon = 1\), suppose that the function class \(\cF\) satisfies~\pref{ass:Bellman_completeness}, and together with the  
underlying MDP admits Bilinear rank $d$. Then with probability at least \(1 - \delta\),  \pref{alg:fqi} obtains the following bound on cumulative subpotimality w.r.t.~any comparator policy \(\pi^e\), 
\begin{align*}
 \sum_{t = 1}^T V^{\pi^e} - V^{\pi^{t}} = \widetilde{O}\prn*{  \max\crl*{C_{\pi^e}, 1} V_{\max} \sqrt{d H^2 T \cdot \log\prn*{ \abs{\cF}/\delta}}},
\end{align*}
where \(\pi^t = \pi^{f^t}\) is the greedy policy w.r.t. \(f^t\) at round \(t\). 
\end{theorem}

The parameters setup in the above theorem indicates that ratio between the total offline samples and the total online sample is each FQI iteration is at least \(1\). This ensures that during learning, we never forget the offline distribution. 
A standard online-to-batch conversion \citep{shalev2014understanding} immediately gives the following sample complexity guarantee for \pref{alg:fqi} for finding an \(\epsilon\)-suboptimal policy w.r.t. the optimal policy \(\pi^*\) for the underlying MDP. 
\begin{corollary}[Sample complexity] 
\label{corr:sample_complexity} 
Under the assumptions of \pref{thm:main_regret} if \(C_{\pi^*} < \infty\) then 
\pref{alg:fqi} can find an \(\epsilon\)-suboptimal policy \(\wh \pi\) for which \(V^{\pi^*} - V^{\wh \pi} \leq \epsilon\) with total sample complexity (online + offline):   
\begin{align*}
n = \widetilde{O}\prn*{ C^2_{\pi^*} V^2_{\max} H^3 d  \log\prn*{\abs{\cF}/\delta}/\epsilon^2} 
\end{align*}
\end{corollary}

The results formalize the statistical properties of Hy-Q. In terms of sample complexity, 
a somewhat unique feature of the hybrid setting is that both transfer coefficient and bilinear rank parameters are relevant, whereas these (or related) parameters typically appear in isolation in offline and online RL respectively. In terms of coverage,~\pref{thm:main_regret} highlights an ``oracle property" of Hy-Q: it competes with \emph{any} policy that is sufficiently covered by the offline dataset.

We also highlight the computational efficiency of Hy-Q: it only requires solving least squares problems over the function class $\Fcal$. To our knowledge, no purely online or purely offline methods are known to be efficient in this sense, except under much stronger ``uniform" coverage conditions. 

\subsection{Proof Sketch}\label{sec:proof_sketch}
We now give an overview of the proof of \pref{thm:main_regret}. The proof starts with a simple decomposition of the regret:
\begin{align*}
    \sum_{t = 1}^T V^{\pi^e} - V^{\pi^{f^t}} &= \sum_{t = 1}^T  \EE_{s \sim d_0}\brk*{V^{\pi^e}_0(s) - V^{\pi^{f^t}}_0(s)} \\ 
    &= \sum_{t = 1}^T \underbrace{\EE_{s \sim d_0}\brk*{V^{\pi^e}_0(s) - \max_a f_0^t(s, a)}}_{A_t} + \sum_{t = 1}^T  \underbrace{\EE_{s \sim d_0}\brk*{ \max_a f_0^t(s, a) - V^{\pi^{f^t}}_0(s)}}_{B_t}.
\end{align*}
Then we note that, one can bound each $A_t$ and $B_t$ by the Bellman error under the comparator $\pi_e$'s visitation distribution and the learned policy's visitation distribution. For simplicity let's define the Bellman error of function $f$ at time $h$ as $\Ecal_h(f)(s,a) = f_h(s,a) - \Tcal f_{h+1}(s,a)$, and we can show that
\begin{align*}
    A_t \leq \sum_{h=0}^{H-1}\EE_{s,a \sim d_h^{\pi^e}}\left[-\Ecal_h(f^t)(s,a)\right] (\text{\pref{lem:optimism}}) \quad\text{and}\quad B_t \leq \sum_{h=0}^{H-1}\left|\EE_{s,a \sim d_h^{\pi^t}}\Ecal_h(f^t)(s,a)\right| (\text{\pref{lem:simulation}}).
\end{align*}
Then for $A_t$, we can recall our definition of the transfer coefficient $C_\pi$ and this gives us
\begin{align*}
    A_t \leq C_{\pi^e} \cdot  \sqrt{ \sum_{h=0}^{H-1} \underbrace{\En_{s, a \sim \nu_h} \brk*{\Ecal_h(f^t)}^2}_{\Ecal_{t;h}^{\textrm{off}}}}.
\end{align*}
The terms $\Ecal^{\textrm{off}}_{t;h}$ are in the order of statistical error resulting from least square regression since every iteration $t$, FQI includes the offline data from $\nu_h$ in its least square regression problems. Thus $A_t$ is small for every $t$ given bounded $C_{\pi^e}$.

To bound the online part $B_t$, we utilize the structure of Bilinear models. For the analysis, we construct a covariance matrix $\Sigma_{t;h} = \sum_{\tau=1}^t X_h(f^\tau)X_h(f^\tau)^{\top} + \lambda \bbI$, where $X_h$ is defined as in the bilinear model construction. This is used to track the online learning progress. Thus recall the definition of bilinear model, we can bound $B_t$ in the following sense:
\begin{align*}
   \sum_{t=1}^T B_t \leq \sum_{t=1}^T\sum_{h=0}^{H-1}\left|\EE_{s,a \sim d_h^{\pi^t}}\Ecal_h(f^t)(s,a)\right| &= \sum_{t=1}^T \sum_{h=0}^{H-1} \abs*{\tri*{X_h(f^t), W_h(f^t)}}\\ &\leq \sum_{t=1}^T \sum_{h=0}^{H-1} \|X_h(f^t)\|_{\Sigma_{t-1; h}^{-1}} \sqrt{\underbrace{\sum_{\tau=1}^{t-1}\EE_{s,a \sim d^\tau_h}[\Ecal_h(f^t)(s,a)]^2}_{\Ecal_{t;h}^{\textrm{on}}} + \lambda B_W^2}.
\end{align*}
The first term on the right hand side of the above inequality, i.e., $\sum_{t} \|X_h(f^t)\|_{\Sigma_{t-1; h}^{-1}}$, can be shown to grow sublinearly $\tilde O(\sqrt{T})$ using the classic elliptical potential argument (\pref{lem:elliptical}).  The term $\Ecal^{\textrm{on}}_{t;h}$ can be controlled to be small as it is related to the statistical error of least square regression (since in each iteration $t$, when we perform least square regression, we use the training data sampled from policies from $\pi^1$ to $\pi^{t-1}$). Together this ensures that $\sum_{t} B_t$ grows sublinearly $\tilde O(\sqrt{T})$, which further implies that there exists a iteration $t'$, such that $B_{t'}  \leq \tilde{O}(1/\sqrt{T})$. Together, the above arguments show that there must exist an iteration $t'$, such that $A_{t'}$ and $B_{t'}$ are small simultaneously, which concludes that $\pi^{f^{t'}}$ can be close to $\pi^e$ in terms of the performance. The proof sketch highlights the key observation: as long as we have a function $f$ that has small Bellman residual under the offline distribution $\nu$ and small Bellman residual under its own greedy policy's distribution $d^{\pi^f}$, then we can show that $\pi^f$ must be at least as good as any policy $\pi^e$ that is covered by the offline distribution.

\subsection{The Linear Bellman Completeness Model}  \label{sec:linear_bellman_complete}
We next showcase one example of low bilinear rank models: the popular linear Bellman complete model which captures both the linear MDP model \citep{yang2019sample, jin2020provably} and the LQR model, and instantiate the sample complexity bound in  \pref{corr:sample_complexity}. 
\begin{definition} 
\label{def:linear_bellman_completeness}
Given a feature function $\phi:\Scal\times\Acal\mapsto \Ball_d(1)$, an MDP with feature $\phi$ admits linear Bellman completeness if for any \(w \in \Ball_d(B_W)\), there exists a \(w' \in \Ball_d(B_W)\) such that 
\begin{align*}
\forall s, a: \qquad \tri*{w', \phi(s,a)} = \EE[R(s,a)] + \EE_{s'\sim \transitionopt(s,a)} \max_{a'} \tri*{w,\phi(s',a')}. 
\end{align*}
\end{definition}

Note that the above condition implies that $Q^\star_h(s,a) = \tri*{w_h^\star,  \phi(s,a)}$ with $\|w^\star_h\|_2 \leq B_W$. Thus, we can define a  function class $\Fcal_h = \{\tri*{w_h, \phi(s,a)}: w_h\in\mathbb{R}^d, \|w_h\|_2 \leq B_W\}$ which by inspection satisfies \pref{ass:Bellman_completeness}. Additionally, this model is also known to have bilinear rank at most $d$ \citep{du2021bilinear}. Thus, using \pref{corr:sample_complexity} we immediately get the following guarantee: \begin{lemma}
\label{lem:linear_Bellman_complete_sc} 
Let \(\delta \in (0, 1)\), suppose the MDP is linear Bellman complete, \(C_{\pi^*} < \infty\), and consider \(\cF_h\) defined above. Then, with probability \(1 - \delta\), \pref{alg:fqi} finds an \(\epsilon\)-suboptimal policy with total sample complexity (offline + online): 
\begin{align*}
n = \widetilde{O}\prn*{\frac{B_W^2 C^2_{\pi^*} H^4 d^2  \log\prn*{B_W/\epsilon\delta}}{\epsilon^2}}.
\end{align*}
\end{lemma}
\begin{proof}[Proof sketch of \pref{lem:linear_Bellman_complete_sc}]
The proof follows by invoking the result in \pref{corr:sample_complexity} for a discretization of the class \(\cF\), denoted by \(\cF_\epsilon = \cF_{0, \epsilon} \times \dots \times \cF_{H-1, \epsilon}\). \(\cF_\epsilon\) is defined such that  $\Fcal_{h, \epsilon} = \{ w^\top \phi(s,a): \widehat \Ball_{d, \epsilon}(B_{W})\}$ where $\widehat  \Ball_{d, \epsilon}(B_{W})$ is an $\epsilon$-net of the $\Ball_{d}(B_{W})$ under \(\ls_\infty\)-distance and contains \(O((B_W/\epsilon)^d)\) many elements. Thus, we get that 
\(\log(\abs*{\cF_\epsilon}) = O\prn*{H d \log(B_W/\epsilon)}\).
\end{proof}

On the computational side, with \(\cF\) as in \pref{lem:linear_Bellman_complete_sc}, the regression problem in \pref{alg:fqi}  reduces to a  \textit{least squares linear} regression with a norm constraint on the weight vector. This can be solved by convex programming with  complexity scaling polynomially in the parameters \citep{bubeck2015convex}.

\begin{remark}[Computational efficiency] 
\label{rem:lbc_remark}
For linear Bellman complete models, we note that  \pref{alg:fqi} can be implemented efficiently under mild assumptions. For the class \(\cF\) in \pref{lem:linear_Bellman_complete_sc}, the regression problem in \pref{eq:alg_regression} reduces to a  \textit{least squares linear} regression with a norm constraint on the weight vector. This regression problem can be solved efficiently by convex programming with  computational efficiency scaling polynomially in the number of parameters \citep{bubeck2015convex} ($d$ here), whenever \(\max_a f_{h+1}(s, a)\) (or $\argmax_a f_{h+1}(s, a)$) can be computed  efficiently.
\end{remark}

\begin{remark}(Linear MDPs)
Since linear Bellman complete models generalize linear MDPs \citep{yang2019sample, jin2020provably}, as we discuss above, \pref{alg:fqi} can be implemented efficiently whenever \(\max_a f_{h+1}(s, a)\) can be computed efficiently. The latter is tractable when: 
\begin{enumerate}[label=\(\bullet\), leftmargin=5mm] 
    \item When \(\abs{\cA}\) is small/finite, one can just enumerate to compute \(\max_a f_{h+1}(s, a)\) for any $s$, and thus \pref{eq:alg_regression} can be implemented efficiently. The computational efficiently of \pref{alg:fqi} in this case is comparable to the prior works, e.g. \cite{jin2020provably}. 
\item When the set \(\crl{\phi(s, a) \mid a \in \cA}\) is convex and compact, one can simply use a linear optimization oracle to compute \(\max_a f_{h+1}(s, a) = \max_{a} w_{h+1}^\top \phi(s, a)\). This linear optimization problem is itself solvable with computational efficiency scaling polynomially with d.  here). 

Note that even under access to a linear optimization oracle, prior works e.g.~ \cite{jin2020provably} rely on  bonuses in the form of $\argmax_{a} \phi(s,a)^\top w + \beta \sqrt{ \phi(s,a)^\top \Sigma \phi(s,a) }$, where $\Sigma$ is some positive definite matrix (e.g., the regularized feature covariance matrix). Computing such bonuses could be NP-hard (in the feature dimension $d$) without additional assumptions \citep{dani2008stochastic}. 
\end{enumerate}
\end{remark}

\begin{remark}(Relative condition number)
A common coverage metric in these linear MDP models is the relative condition number. In~\pref{app:coef}, we show that our coefficient $C_\pi$ is upper bounded by the relative condition number of $\pi$ with respect to $\nu$: $\EE_{d^{\pi}}\|\phi\|_{\Sigma^{-1}_{\nu}}$, where $\Sigma_{\nu} = \EE_{s,a \sim \nu}\phi(s,a)\phi^{\top}(s,a)$. Concretely, we have $C_\pi \leq \sqrt{\max_h\EE_{d^{\pi}_h}\|\phi\|^2_{\Sigma^{-1}_{\nu_h}}}$. Note that such quantity captures coverage in terms of feature, and can be bounded even when density ratio style concentrability coefficient (i.e., $\sup_{s,a} d^{\pi}(s,a) / \nu(s,a)$) being infinite.
\end{remark}

\subsection{Low-rank MDP} 
In this section, we briefly introduce the low-rank MDP  model \citep{du2021bilinear}, which is captured by the V-type Bilinear model discussed in \pref{app:V_type_appendix}. 
 Unlike the linear MDP model discussed in \pref{sec:linear_bellman_complete}, low-rank MDP does not assume the feature $\phi$ is known a priori. 
\begin{definition}[Low-rank MDP]
\label{def:low_rank_mdp}
A MDP is called low-rank MDP if there exists $\mu^\star:\Scal\mapsto \mathbb{R}^d, \phi^\star:\Scal\times\Acal\mapsto \mathbb{R}^d$, such that the transition dynamics $\transitionopt(s' | s,a) = \mu^\star(s')^\top \phi^\star(s,a)$ for all $s,a,s'$. We additionally assume that we are given a realizable representation class $\Phi$ such that $\phi^\star \in \Phi$, and that $\sup_{s,a} \| \phi^\star(s,a) \|_2 \leq 1$, and $\| f^\top \mu^\star \|_2 \leq \sqrt{d}$ for any $f:\Scal\mapsto [-1,1]$.
\end{definition}

Consider the function class $\Fcal_h = \{ w^\top \phi(s,a): \phi\in \Phi, w \in \Ball_d(B_{W})\}$, and through the bilinear decomposition we have that $B_W \leq 2\sqrt{d}$.  
By inspection, we know that this function class satisfies \pref{ass:Bellman_completeness}. Furthermore, it is well known that the low rank MDP model has V-type bilinear rank of at most \(d\)
 \citep{du2021bilinear}. Invoking the sample complexity bound given in \pref{corr:Vtype_sample_complexity} for V-type Bilinear models, we get  the following result. 
\begin{lemma}
\label{lem:low_rank_MDP}
Let \(\delta \in (0, 1)\) and \(\Phi\) be a given representation class. Suppose that the MDP is a rank \(d\) MDP w.r.t. some \(\phi^\star \in \Phi\), \(C_{\pi^*} < \infty\), and consider \(\cF_h\) defined above. Then, with probability \(1 - \delta\), \pref{alg:vtype} finds an \(\epsilon\)-suboptimal policy with total sample complexity (offline + online): 
\begin{align*}
    \wt{O} \prn*{ \frac{\max\crl*{C^2_{\pi^*}, 1} \Vmax^2 d^2  H^4 \abs*{\cA} \log\prn*{{HT d\abs{\Phi}}/{\epsilon \delta}}}{\epsilon^2}}. 
\end{align*} 
\end{lemma} 

\begin{proof}[Proof sketch of \pref{lem:low_rank_MDP}]
The proof follows by invoking the result in \pref{corr:sample_complexity} for a discretization of the class \(\cF\), denoted by \(\cF_\epsilon = \cF_{0, \epsilon} \times \dots \times \cF_{H-1, \epsilon}\). \(\cF_\epsilon\) is defined such that $\Fcal_{h, \epsilon} = \{ w^\top \phi(s,a): \phi\in \Phi, w \in \widehat \Ball_{d, \epsilon}(B_{W})\}$ where $\widehat  \Ball_{d, \epsilon}(B_{W})$ is an $\epsilon$-net of the $\Ball_{d}(B_{W})$ under \(\ls_\infty\)-distance and contains \(O((B_W/\epsilon)^d)\) many elements. Thus, we get that 
\(\log(\abs*{\cF_\epsilon}) = O\prn*{H d \log(B_W \abs*{\Phi}/\epsilon)}\). 
\end{proof}

For low-rank MDP, the transfer coefficient $C_{\pi}$ is upper bounded by a relative condition number style quantity defined using the unknown ground truth feature $\phi^\star$ (see \pref{lem:low_rank_mdp_transfer}). 
On the computational side, \pref{alg:fqi} (with the modification of $a\sim \text{Uniform}(\Acal)$ in the online data collection step) requires to solve a least squares regression problem at every round. The objective of this regression problem is a convex functional of the hypothesis \(f\) over the constraint set \(\cF\). While this is not fully efficiently implementable due to the potential non-convex constraint set \(\cF\) (e.g., $\phi$ could be complicated), our regression problem is still much simpler than the oracle models considered in the prior works for this model \citep{agarwal2020flambe,sekhari2021agnostic,uehara2021representation,modi2021model}.

\begin{figure}
  \begin{center} 
    \includegraphics[height=0.3\textwidth]{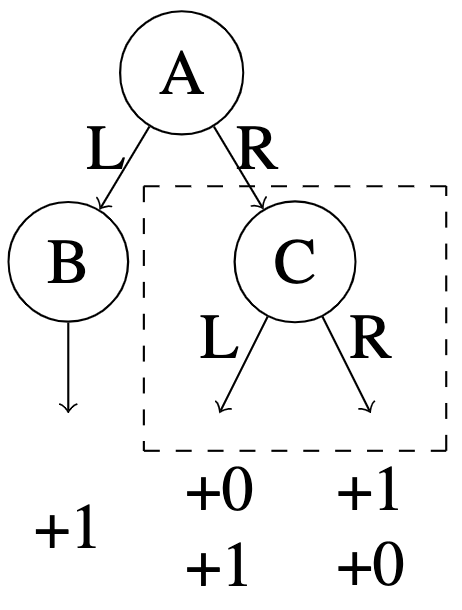}
         \caption{A hard instance for offline RL \citep[reproduced with permission]{zhan2022offline}}
   \label{fig:lower_bound} 
  \end{center} 
\end{figure}

\subsection{Why don't offline RL methods work?} 
One may wonder why do pure offline RL methods fail to learn when the transfer coefficient is bounded, and why does online access help? We illustrate with the MDP construction developed by~\citet{zhan2022offline,chen2022offline}, visualized in~\pref{fig:lower_bound}.

Consider two  MDPs \(\crl*{M_1, M_2}\) with \(H=2\), three states \(\crl*{A, B, C}\), two actions \(\crl*{L, R}\) and the fixed start state \(A\). The two MDPs have the same dynamics but different rewards. In both, actions from state $B$ yield reward $1$. In $M_1$, $(C,R)$ yields reward $1$ while $(C,L)$ yields reward $1$ in $M_2$. 
All other rewards are \(0\). In both \(M_1\) and \(M_2\), an optimal policy is  \(\pi^*(A) = L\) and \(\pi^*(B)=\pi^*(C)=\mathrm{Uniform}\prn*{\crl*{L, R}}\). 
With $\Fcal = \{Q_1^\star, Q_2^\star\}$ where $Q_j^\star$ is the  optimal $Q$ function for $M_j$, then one can easily verify that $\Fcal$ satisfies Bellman completeness, for both MDPs. Finally with offline distribution $\nu$ supported on states $A$ and $B$ only (with no coverage on state $C$), we have sufficient coverage over $d^{\pi^\star}$. However, samples from $\nu$ are unable to distinguish between $f_1$ and $f_2$ or ($M_1$ and $M_2$), since state $C$ is not supported by $\nu$. Unfortunately,  adversarial tie-breaking may result the greedy policies of $f_1$ and $f_2$ visiting state $C$, where we have no information about the correct action. \looseness=-1

This issue has been documented before, and in order to address it with pure offline RL, existing approaches require additional structural assumptions. For instance, \cite{chen2022offline} assume that $Q^\star$ has a gap, which usually does not hold when action space is large or continuous.  \cite{xie2021bellman} assumes policy-dependent Bellman completeness for every possible policy $\pi\in \Pi$ (which is much stronger than our assumption),
and \cite{zhan2022offline} assumes a somewhat non-interpretable realizability assumption on some ``value" function that does not obey the standard Bellman equation. In contrast, by combining offline data and online data, our approach focuses on functions that have small Bellman residual under both the offline distribution and the on-policy distributions, which  together with the offline data coverage assumption, ensures near optimality. It is easy to see that the hybrid approach will succeed~\pref{fig:lower_bound}.

\section{Experiments}

In this section we discuss empirical results comparing Hy-Q to several representative RL methods on two challenging benchmarks. Our experiments focus on answering the following questions:
\begin{enumerate}
    \item Can Hy-Q efficiently solve problems that SOTA offline RL methods simply cannot?
    \item Can Hy-Q, via the use of offline data, significantly improve the sample efficiency of online RL? 
    \item Does Hy-Q scale to challenging deep-RL benchmarks?
\end{enumerate}

\begin{figure}
  \begin{center}
    \includegraphics[width=0.45\textwidth]{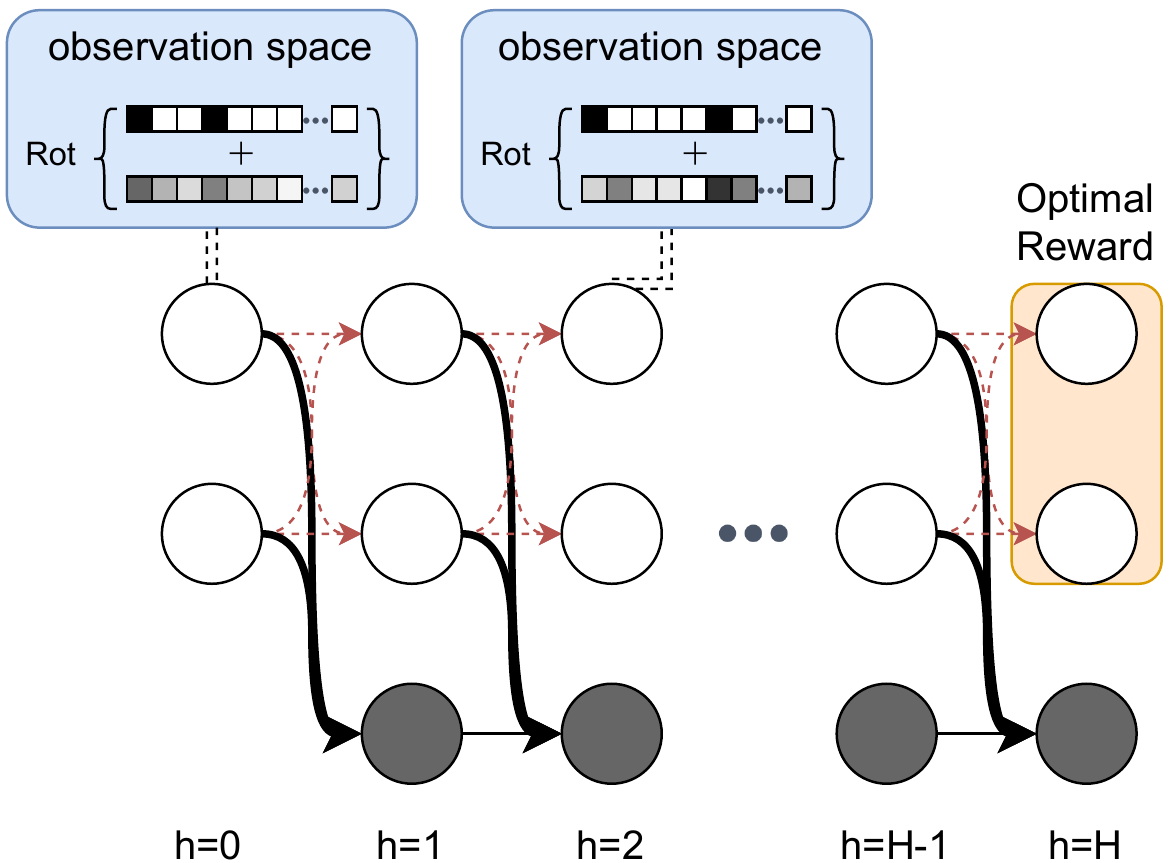}
         \caption{The rich observation combination lock \citep{misra2020kinematic,zhang2022efficient}.} 
   \label{fig:combination lock diagram} 
  \end{center}
\end{figure}

Our empirical results provide positive answers to all of these questions. To study the first two, we consider the diabolical combination lock environment~\citep{misra2020kinematic,zhang2022efficient}, a synthetic environment designed to be particularly challenging for online exploration. The synthetic nature allows us to carefully control the offline data distribution to modulate the difficulty of the setup and also to compare with a provably efficient baseline~\citep{zhang2022efficient}. To study the third question, we consider the Montezuma's Revenge benchmark from the 
Arcade Learning environment, which is one of the most challenging empirical benchmarks with high-dimensional image inputs, largely due to the difficulties of exploration. 
Additional details are deferred to~\pref{app:exp}.

\paragraph{Hy-Q implementation.} We largely follow \pref{alg:fqi} in our implementation for the combination lock experiment. Particularly,  we use a similar function approximation to \citet{zhang2022efficient}, and a minibatch Adam update on Eq.~\pref{eq:alg_regression} with the same sampling proportions as in the pseudocode. For Montezuma's Revenge, in addition to minibatch optimization, since the horizon of the environment is not fixed, we deploy a discounted version of Hy-Q. Concretely, the target value in the Bellman error is calculated 
from  the output of a target network, which is periodically updated, times a discount factor. We refer the readers to~\pref{app:exp} for more details.  

\paragraph{Baselines.} We include representative algorithms from four categories: (1) for imitation learning we use Behavior Cloning (\textsc{Bc}) \citep{bain1995framework}, (2) for offline RL we use Conservative Q-Learning (\textsc{Cql})~\citep{kumar2020conservative} due to its successful demonstrations on some Atari games, (3) for online RL we use  \textsc{Briee}~\citep{zhang2022efficient} for combination lock\footnote{We note that \textsc{Briee} is currently the state-of-the-art method for the combination lock environment. In particular,~\citet{misra2020kinematic} show that many Deep RL baselines fail in this environment.} and Random Network Distillation (\textsc{Rnd})~\citep{burda2018exploration} for Montezuma's Revenge, and (4) as a Hybrid-RL baseline we use Deep Q-learning from Demonstrations (\textsc{Dqfd})~\citep{hester2018deep}. We note that \textsc{Dqfd} and prior hybrid RL methods combine expert demonstrations with online interactions, but are not necessarily designed to work with general offline datasets.

\paragraph{Results summary.} Overall, we find that Hy-Q performs favorably against all of these baselines. Compared with offline RL, imitation learning, and prior hybrid methods, Hy-Q is significantly more robust in the presence of a low quality offline data distribution. Compared with online methods, Hy-Q offers order-of-magnitude savings in the total experience. 

\paragraph{Reproducibility.} We release our code at \url{https://github.com/yudasong/HyQ}. We also include implementation details in \pref{app:exp}. 

\begin{figure} 
    \centering
    \includegraphics[width=1.0\linewidth]{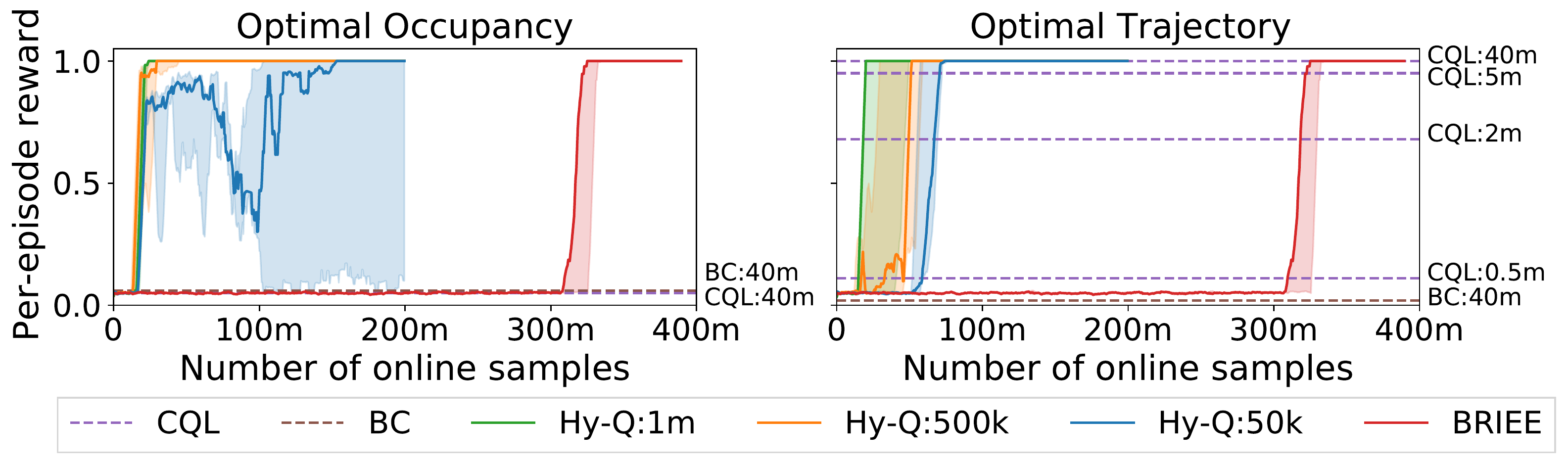}
    \caption{The learning curve for combination lock with $H=100$. The plots show the median and 80th/20th quantile for 5 replicates. Pure offline and IL methods are visualized as dashed horizontal lines (in the left plot, CQL overlaps with BC). Note that we report the number of samples while~\citet{zhang2022efficient} report the number of episodes.}
    \label{fig:combination lock}
\end{figure}

\subsection{Combination Lock}

The combination lock benchmark is depicted in~\pref{fig:combination lock diagram} and consists of horizon $H=100$, three latent states for each time step and $10$ actions in each state. Each state has a single ``good" action that advances down a chain of favorable latent states (white) from which optimal reward can be obtained. A single incorrect action transitions to an absorbing chain (black latent states) with suboptimal value. The agent operates on high dimensional continuous observations omitted from latent states and must use function approximation to succeed. This is an extremely challenging problem for which many Deep RL methods are known to fail~\citep{misra2020kinematic}, in part because (uniform) random exploration only has $10^{-H}$ probability of obtaining the optimal reward.

On the other hand, the model has low bilinear rank, so we do have online RL algorithms that are provably sample-efficient: \textsc{Briee} \citep{zhang2022efficient} currently obtains state of the art sample complexity. However, its sample complexity is still quite large, and we hope that Hybrid RL can address this shortcoming. We are not aware of any experiments with offline RL methods on this benchmark. 

We construct two offline datasets for the experiments, both of which are derived from the optimal policy $\pi^\star$ which always picks the "good" actions and stays in the chains of white states. In the \textbf{optimal trajectory} dataset we collect full trajectories by following $\pi^\star$ with $\epsilon$-greedy exploration with $\epsilon=1/H$. We also add some noise by making the agent to perform randomly at timestep $H/2$.
In the \textbf{optimal occupancy} dataset we collect transition tuples from the state-occupancy measure of $\pi^\star$ with random actions.\footnote{Formally,  we sample $h \sim \textrm{Unif}([H])$, $s \sim d_h^{\pi^\star}$, $a \sim \textrm{Unif}(\Acal)$, $r \sim R(s,a)$, $s' \sim P(s,a)$.} Both datasets have bounded concentrability coefficients (and hence transfer coefficients) with respect to $\pi^\star$, but the second dataset is much more challenging since the actions in the offline dataset do not directly provide information about $\pi^\star$, as they do in the former. 

The results are presented in~\pref{fig:combination lock}. First, we observe that Hy-Q can reliably solve the task under both offline distributions with relatively low sample complexity (500k offline samples and $\leq$ 25m online samples). In comparison, \textsc{Bc} fails completely since both datasets contain random actions. \textsc{Cql} can solve the task using the  optimal trajectory-based dataset with a sample complexity that is comparable to the combined sample size of Hy-Q. However, \textsc{Cql} fails on the optimal occupancy-based dataset since the actions themselves are not informative. Indeed the pessimism-inducing regularizer of \cql is constant on this dataset and so the algorithm reduces to \textsc{Fqi} which provably fails when the offline data does not have a global coverage (i.e., covers every state-action pair). Finally, Hy-Q can solve the task with a factor of 5-10 reduction in samples (online plus offline) when compared with \textsc{Briee}. This demonstrates the robustness and sample efficiency provided by hybrid RL.

\begin{figure} 
    \centering
    \includegraphics[width=1\linewidth]{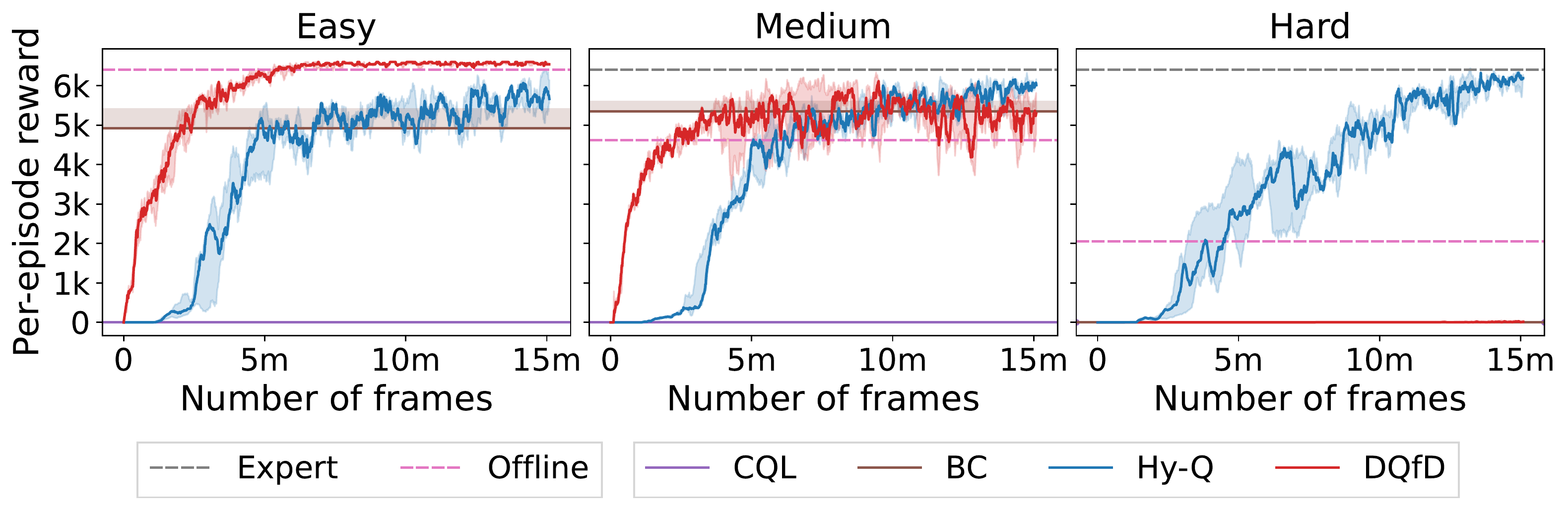}
    \caption{The learning curve for Montezuma's Revenge.  The plots show the median and 80th/20th quantile for 5 replicates. Pure offline, IL methods and dataset qualities are visualized as dashed horizontal lines. ``Expert'' denotes $V^{\pi^e}$ and ``Offline'' denotes the average trajectory reward in the offline dataset. The y-axis denotes the (moving) average of 100 episodes for the methods involving online interactions. Note that \cql and \textsc{Bc} overlap on the last plot. \vspace{-0.3cm}}
    \label{fig:mrevenge}
\end{figure}

\begin{figure}
\centering 
\includegraphics[width=1\linewidth]{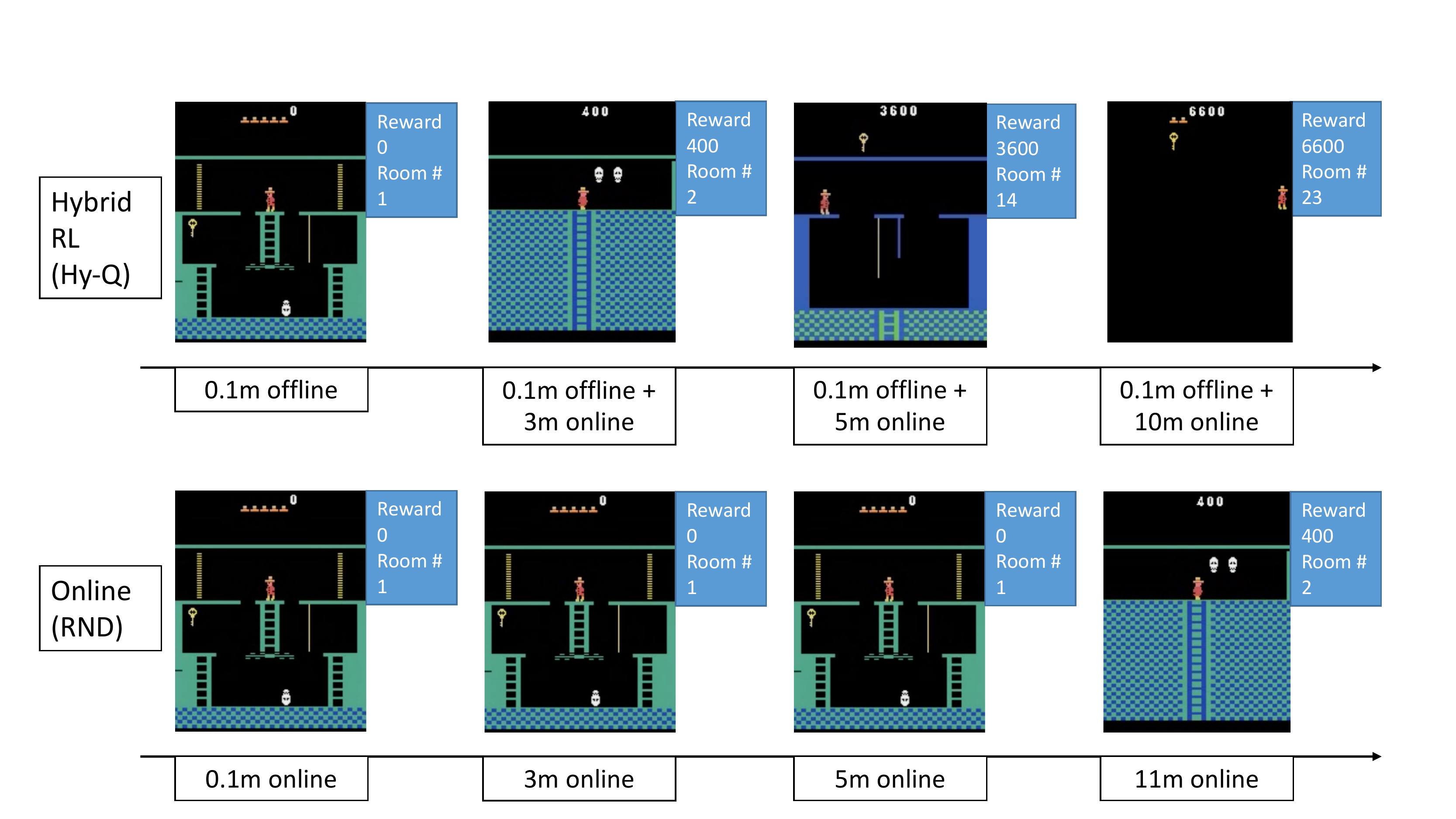} \vspace{5pt}
\caption{Screenshots of the training processes of our approach \algname{} (top row) and \textsc{Rnd} (bottom row). With  only 0.1m offline samples of which half is from a random policy and half is from a high quality policy (with reward around 6400), our approach learns significantly faster than \textsc{Rnd}.}
\label{fig:mrevenge_visual}
\end{figure}

\subsection{Montezuma's Revenge} 
\label{sec:m_revenge}

To answer the third question, we turn to Montezuma's Revenge, an extremely challenging image-based benchmark environment with sparse rewards. We follow the setup from~\citet{burda2018exploration} and introduce stochasticity to the original dynamics: with probability 0.25 the environment executes the previous action instead of the current one. For offline datasets, we first train an ``expert policy" $\pi^e$ via \textsc{Rnd} to achieve $V^{\pi^e} \approx 6400$. We create three datasets by mixing samples from $\pi^e$ with those from a random policy: the \textbf{easy dataset} contains only samples from $\pi^e$, the \textbf{medium dataset} mixes in a 80/20 proportion (80 from $\pi^e$), and the \textbf{hard dataset} mixes in a 50/50 proportion. Here we record full trajectories from both policies in the offline dataset, but measure the proportion using the number of transition tuples instead of trajectories. We provide 0.1 million offline samples for the hybrid methods, and 1 million samples for the offline and IL methods.

Results are displayed in~\pref{fig:mrevenge}. \textsc{Cql} fails completely on all datasets. \textsc{Dqfd} performs well on the easy dataset due to the supervised learning style large margin loss \citep{piot2014boosted} that imitates the policies in the offline dataset. However, \dqfd's performance drops as the quality of the offline dataset degrades (medium), and fails when the offline dataset is low quality (hard), where one cannot simply treat offline samples as expert samples. We also observe that \textsc{Bc} is a competitive baseline in the first two settings due to high fraction of expert samples, and thus we view these problems as relatively easy to solve. Hy-Q is the only method that performs well on the hard dataset. Note that here, BC's performance is quite poor.  
We also include the comparison with \rnd in \pref{fig:rnd} and \pref{fig:mrevenge_visual}: with only 100k offline samples 
from any of the three datasets,  Hy-Q is over 10x more efficient in terms of online sample complexity. \looseness=-1

\section{Conclusion} 

We demonstrate the potential of hybrid RL with Hy-Q, a simple, theoretically principled, and empirically effective algorithm. Our theoretical results showcase how Hy-Q circumvents the computational issues of pure offline or online RL, while our empirical results highlight its robustness and sample efficiency. Yet, Hy-Q is perhaps the most natural hybrid algorithm, and we are optimistic that there is much more potential to unlock from the hybrid setting. We look forward to studying this in the future. 

\section*{Acknowledgement} 
AS thanks Karthik Sridharan for useful discussions. WS acknowledges funding support from NSF IIS-2154711. We thank Simon Zhai for their careful reading of the manuscript and improvement on the technical correctness of our paper. We also thank Uri Sherman for their discussion on the computational efficiency of the original draft.

\setlength{\bibsep}{6pt} 
\bibliography{reference,ref}

\clearpage 

\setlength\parindent{0pt}
\setlength{\parskip}{0.25em} 

\appendix
\section{Proofs for \pref{sec:low_bilinear_rank}}

\paragraph{Additional notation.}  Throughout the appendix, we define the feature covariance matrix \(\Sigma_{t; h}\) as 
\begin{align*}
    \Sigma_{t;h} = \sum_{\tau = 1}^t X_h(f^\tau) \prn*{X_h(f^\tau)}^\top + \lambda \bbI.  \numberthis \label{eq:sigma_defn} 
\end{align*}

Furthermore, given a distribution $\beta \in \Delta(\Scal\times\Acal)$ and a function $f:\Scal\times\Acal\mapsto \mathbb{R}$, we denote its weighted $\ell_2$ norm as $\| f \|^2_{2,\beta} := \sqrt{ \EE_{s,a\sim \beta} f^2(s,a) }$. 

\subsection{Supporting lemmas for \pref{thm:main_regret}}
Before proving \pref{thm:main_regret}, we first present a few useful lemma. We start with a standard result on least square generalization bound, which is be used by recalling that \pref{alg:fqi} performs least squares on the empirical bellman error. We defer the proof of \pref{lem:sq_loss_generalization} to \pref{app:support}.
\begin{lemma}(Least squares generalization bound)  
\label{lem:sq_loss_generalization} 
Let \(R > 0\), \(\delta \in (0, 1)\), we consider a sequential function estimation setting, with an instance space $\Xcal$ and target space $\Ycal$. Let \(\cH: \cX \mapsto [-R, R]\) be a class of real valued functions. Let \(\cD = \crl*{(x_1, y_1), \dots, (x_T, y_T)}\) be a dataset of \(T\) points where $x_t \sim \distrib_t \ldef{} \distrib_t(x_{1:t-1},y_{1:t-1})$, and \(y_t\) is sampled via the conditional probability $p(\cdot \mid x_t)$:
\begin{align*}
y_t \sim  p(\cdot \mid x_t) \ldef{} h^*(x_t) + \varepsilon_t, 
\end{align*} where the function \(h^*\) satisfies approximate realizability i.e. $$\inf_{h \in \cH}  \frac{1}{T} \sum_{t=1}^T  \En_{x \sim \distrib_t} \brk*{\prn*{h^*(x) - h(x)}^2} \leq \gamma,$$ and \(\crl*{\epsilon_i}_{i=1}^n\)  are independent  random variables such that $\EE[y_t \mid x_t] = h^{\ast}(x_t)$. Additionally, suppose that \(\max_t \abs{y_t} \leq R\) and \(\max_{x} \abs*{h^*(x)} \leq R\). Then the least square solution \(\wh h \leftarrow \argmin_{h \in \cH} \sum_{t=1}^T \prn*{h(x_t) - y_t}^2\) satisfies with probability at least \(1 - \delta\), 
\begin{align*}
 \sum_{t=1}^T \En_{x \sim \distrib_t} \brk*{\prn{\wh h(x) - h^*(x)}^2} &\leq 3 \gamma T + 256 R^2 \log(2 \abs{\cH}/\delta). 
\end{align*}
\end{lemma} 
The above lemma is basically an extension of the standard least square regression  agnostic generalization bound from i.i.d.~setting to the non-i.i.d.~case with the sequence of training data forms a sequence of Martingales. We state the result when the realizability only holds approximately upto the approximation \(\gamma\). However, for all our proofs, we invoke this result by setting \(\gamma = 0\). 

In the next two lemmas, we prove two lemmas where we can bound each part of the regret decomposition using the Bellman error of the value function $f$.
\begin{lemma}[Performance difference lemma]
\label{lem:simulation}
For any function \(f = \prn*{f_0, \dots, f_{H-1}}\) where \(f_h: \cS \times \cA \mapsto \bbR\) and \(h \in [H-1]\), we have 
\begin{align*}
    \EE_{s \sim d_0}[\max_a f_0(s,a)- V_0^{\pi^{f}}(s)] \le \sum_{h = 0}^{H - 1} \abs*{ \EE_{s,a \sim d_h^{\pi^{f}}}\brk*{f_h(s,a) - \dynamicsopt f_{h+1}(s,a)}}, 
\end{align*}
where we define \(f_H(s, a) = 0\) for all \(s, a\). 
\end{lemma} 
\begin{proof} We start the proof by noting that $\pi^f_0(s) = \argmax_a f_0(s,a)$, then we have: 
\begin{align*}
    \EE_{s \sim d_0}[\max_a f_0(s,a)- V^{\pi^{f}}(s)] &=  \EE_{s \sim d_0}[\EE_{a \sim \pi_0^f(s)}f_0(s,a) - V_0^{\pi^f}(s)]\\
    &=  \EE_{s \sim d_0}[\EE_{a \sim \pi_0^f(s)}f_0(s,a) -\dynamicsopt f_1(s,a)] + \EE_{s \sim d_0}[\EE_{a \sim \pi^f_0(s)}\dynamicsopt f_1(s,a) - V_0^{\pi^f}(s)]\\
    &= \EE_{s,a \sim d^{\pi^f}_0}[f_0(s,a) -\dynamicsopt f_1(s,a)] + \\ &\;\;\;\; \EE_{s \sim d_0}[\EE_{a \sim \pi_0^f(s)}[R(s,a) + \gamma \EE_{s' \sim \Pcal(s,a)} \max_{a'}f_1(s',a') - R(s,a) +  \EE_{s' \sim \Pcal(s,a)} V^{\pi^f}_1(s')]]\\
    &= \EE_{s,a \sim d^{\pi^f}_0}[f_0(s,a) -\dynamicsopt f_1(s,a)] +  \EE_{s \sim d^{\pi^f}_1}[\max_a f_1(s,a) - V_1^{\pi^f}(s)]
    \numberthis \label{eq:simulation_lemma_1} 
\end{align*}
Then by recursively applying the same procedure on the second term in \pref{eq:simulation_lemma_1}, we have 
\begin{align*}
     \EE_{s \sim d_0}[\max_a f_0(s,a)- V^{\pi^{f}}(s)] &= \sum_{h=0}^{H-1} \EE_{s,a \sim d^{\pi^f}_h}[f_h(s,a) -\dynamicsopt f_{h+1}(s,a)] + \EE_{s \sim d^{\pi^f}_{H}}[\max_a f_H(s,a) - V_H^{\pi^f}(s)].
\end{align*}
Finally for $h = H$, we recall that we set $f_{H}(s,a) = 0$ and $V^{\pi^f}_H = 0$ for notation simplicity. Thus we have:
\begin{align*}
     \EE_{s \sim d_0}[\max_a f_0(s,a)- V^{\pi^{f}}(s)] &= \sum_{h=0}^{H-1} \EE_{s,a \sim d^{\pi^f}_h}[f_h(s,a) -\dynamicsopt f_{h+1}(s,a)]\\
     &\leq \sum_{h=0}^{H-1} \left|\EE_{s,a \sim d^{\pi^f}_h}[f_h(s,a) -\dynamicsopt f_{h+1}(s,a)]\right|.
\end{align*}
\end{proof}

Now we proceed to how to bound the other half in the regret decomposition:
\begin{lemma}
\label{lem:optimism}
Let \(\pi^e = \prn{\pi^e_0, \dots, \pi^e_{H-1}}\) be a comparator policy, and consider any value function \(f = \prn{f_0, \dots, f_{H-1}}\) where \(f_h: \cS \times \cA \mapsto \bbR\). Then, 
    \begin{align*}
\EE_{s \sim d_0}  \brk*{ V^{\pi^e}_0(s) - \max_a f_0(s, a)} \leq \sum_{i = 0}^{H - 1}\EE_{s,a \sim d^{\pi_e}_i}[\dynamicsopt  f_{i+1}(s,a) - f_i(s,a)], 
    \end{align*} 
   where we defined \(f_H(s, a) = 0\)  for all \(s, a\). 
\end{lemma} 
\begin{proof} The proof is similar to the proof of \pref{lem:simulation}, and we start with the fact that $\max_a f(s,a) \geq f(s,a'), \forall a'$, including actions sampled from $\pi^e$:
\begin{align*}
\EE_{s \sim d_0}  \brk*{ V_0^{\pi^e}(s) - \max_a f_0(s, a)} &\leq \EE_{s,a \sim d^{\pi_e}_0} \brk*{Q_0^{\pi^e}(s,a) - f_0(s,a)}\\
&= \EE_{s,a \sim d^{\pi_e}_0} \brk*{Q_0^{\pi^e}(s,a) - \Tcal f_1(s,a) + \Tcal f_1(s,a) - f_0(s,a)}\\
&= \EE_{s,a \sim d^{\pi_e}_0} \brk*{  \EE_{s' \sim \Pcal(s,a)}V_1^{\pi^e}(s') - \max_{a'} f_1(s',a')} + \EE_{s,a \sim d^{\pi_e}_0}\brk*{ \Tcal f_1(s,a) - f_0(s,a)}\\
&= \EE_{s\sim d^{\pi_e}_1} \brk*{ V_1^{\pi^e}(s) - \max_{a} f_1(s,a)} + \EE_{s,a \sim d^{\pi_e}_0}\brk*{ \Tcal f_1(s,a) - f_0(s,a)} \numberthis \label{eq:optimism1}
\end{align*}
Again by recursively applying the same procedure on  the first term in \pref{eq:optimism1}, we have
\begin{align*}
    \EE_{s \sim d_0}  \brk*{ V_0^{\pi^e}(s) - \max_a f_0(s, a)} &\leq \EE_{s\sim d^{\pi_e}_H} \brk*{ V_H^{\pi^e}(s) - \max_{a} f_H(s,a)} + \sum_{h=0}^{H-1} \EE_{s,a \sim d^{\pi_e}_h}\brk*{ \Tcal f_{h+1}(s,a) - f_h(s,a)},
\end{align*}
and recall that $f_{H}(s,a) = 0$ and $V^{\pi^f}_H = 0$, we have
\begin{align*}
    \EE_{s \sim d_0}  \brk*{ V_0^{\pi^e}(s) - \max_a f_0(s, a)} &\leq  \sum_{h=0}^{H-1} \EE_{s,a \sim d^{\pi_e}_h}\brk*{ \Tcal f_{h+1}(s,a) - f_h(s,a)}.
\end{align*}
\end{proof}

The following result is useful in the bilinear models when we want to bound the potential functions. The result directly follows from the elliptical potential lemma \citep[Lemma 19.4]{lattimore2020bandit}. 
\begin{lemma}  
\label{lem:elliptical} 
Let \(X_h(f^1), \dots, X_h(f^T) \in \bbR^d\) be a sequence of vectors with \(\nrm{X_h(f^t)} \leq B_X < \infty\) for all \(t \leq T\). Then, 
\begin{align*}
    \sum_{t=1}^T \|X_h(f^t)\|_{\Sigma_{t - 1;h}^{-1}} \leq \sqrt{2 d T \log\prn*{1 + \frac{T B_X^2}{\lambda d}}}, 
\end{align*} 
where the matrix $\Sigma_{t;h} \ldef{} \sum_{\tau = 1}^t X_h(f^\tau) X_h(f^\tau)^\top + \lambda \bbI$~for \(t \in [T]\) and \(\lambda \geq B^2_X\), and the matrix norm $\|X_h(f^t)\|_{\Sigma_{t-1;h}^{-1}} = \EE[X_h(f^t)^{\top} \Sigma_{t-1;h}^{-1} X_h(f^t)]$. 
\end{lemma}
\begin{proof} Since \(\lambda \geq B_X^2\), we have that 
\begin{align*} 
\nrm{X_h(f^t)}^2_{\Sigma^{-1}_{t - 1; h}} \leq \frac{1}{\lambda} \nrm{X_h(f^t)}^2 \leq 1. 
\end{align*} 
Thus, using elliptical potential lemma \citep[Lemma 19.4]{lattimore2020bandit}, we get that 
\begin{align*}
    \sum_{t=1}^T \|X_h(f^t)\|^2_{\Sigma_{t - 1;h}^{-1}} \leq 2 d\log\prn*{1 + \frac{TB_X^2}{\lambda d}}. 
\end{align*}  
	
The desired bound follows from Jensen's inequality which implies that 
\begin{align*}
    \sum_{t=1}^T \|X_h(f^t)\|_{\Sigma_{t - 1;h}^{-1}}  \leq \sqrt{T \cdot \sum_{t=1}^T \|X_h(f^t)\|^2_{\Sigma_{t - 1;h}^{-1}}} \leq \sqrt{2 T d \log\prn*{1 + \frac{TB_X^2}{\lambda d}}}. 
\end{align*} 
\end{proof}

\subsection{Proof of \pref{thm:main_regret}} \label{sec:main_proofs}
Before delving into the proof, we first state that following generalization bound for FQI.
\begin{lemma}[Bellman error bound for FQI]
\label{lem:fqi_guarantee} 
Let \(\delta \in (0,1)\) and let for \(h \in [H-1]\) and \(t \in [T]\), \(f^{t+1}_h\) be the estimated value function for time step \(h\) computed via least square regression using samples in the dataset \({\prn*{\cD^\nu_h, \cD^{1}_h, \dots, \cD^{t}_h}}\) in \pref{eq:alg_regression} in the iteration \(t\) of \pref{alg:fqi}. Then, with probability at least \(1 - \delta\), for any \(h \in [H-1]\) and \(t \in [T]\), 
\begin{align*}
\nrm*{f_h^{t+1} - \dynamicsopt f_{h+1}^{t+1}}_{2, \nu_h}^2 \leq   \frac{1}{\moff} 256 \Vmax^2 \log(2 H T \abs{\cF}/\delta)  \rdef{} \Delta_{\mathrm{off}},
\intertext{and}
\sum_{\tau = 1}^t  \nrm*{f_h^{t+1} - \dynamicsopt f_{h+1}^{t+1}}^2_{2, \mu^\tau_h} \leq \frac{1}{\mon}  256 \Vmax^2 \log(2 H T \abs{\cF}/\delta)  \rdef{} \Delta_{\mathrm{on}},
\end{align*} where \(\nu_h\) denotes the offline data distribution at time \(h\), and the distribution \(\mu^\tau_h \in \Delta(s, a)\) is defined such that \(s, a \sim d^{\pi^\tau}_h\). 
\end{lemma}
\begin{proof} Fix \(t \in [T]\), \(h \in [H-1]\) and \(f^{t+1}_{h+1} \in \cF_{h+1}\) and consider the regression problem (\pref{eq:alg_regression} in the iteration \(t\) of \pref{alg:fqi}):
 \begin{align*}
f_{h}^{t+1} \leftarrow \argmin_{f\in \Fcal_h}   \crl*{
 \moff \wh \EE_{\Dcal^\nu_{h}}( f(s,a) -   r -  \max_{a'} f^{t+1}_{h+1}(s', a') )^2   + \mon \sum_{\tau = 1}^{t} \wh \EE_{\Dcal^\tau_{h}}( f(s,a) -   r -  \max_{a'} f^{t+1}_{h+1}(s', a') )^2} ,
\end{align*}
which can be thought of as regression problem
 \begin{align*}
f_{h}^{t+1} \leftarrow \argmin_{f\in \Fcal_h}   \crl*{
\wh \EE_{\Dcal}( f(s,a) -   r -  \max_{a'} f^{t+1}_{h+1}(s', a') )^2 }, 
\end{align*}
where dataset \(\cD\) consisting of \(n = \moff + t \cdot \mon\) samples \(\crl*{(x_i, y_i)}_{i \leq n}\) where 
\begin{align*}
x_i = (s^i_h, a^i_h) \qquad \text{and} \qquad y^i = r^i + \max_{a} f_{h+1}^{t+1}(s^i_{h+1}, a). 
\end{align*}
In particular, we define \(\cD\) such that the first \(\moff\) samples \(\crl*{(x_i, y_i)}_{i \leq \moff} = \cD^{\nu}_h\),  the next \(\mon\) samples  \(\crl*{(x_i, y_i)}_{i = \moff + 1}^{\moff + \mon} = \cD^1_h\), and so on where the samples \(\crl*{(x_i, y_i)}_{i = \moff + (\tau - 1) \mon + 1}^{\moff + \tau \mon} = \cD^{\tau}_h\). Note that: (a) for any sample \((x = (s_h, a_h), y = (r + \max_{a} f_{h+1}^{t+1}(s_{h+1}, a)))\) in \(\cD\), we have that 
\begin{align*}
\En \brk*{y \mid x} &= \En_{s_{h+1} \sim \transitionopt(s_h, a_h), r \sim R(s_h, a_h)} \brk*{r + \max_{a} f_{h+1}^{t+1}(s_{h+1}, a)} \\
&= \dynamicsopt f_{h+1}^{t+1}(s_h, a_h) \leq g(s_h, a_h), 
\end{align*} where the last line holds since the Bellman completeness assumption implies existence of such a function \(g\), (b) for any sample, \(\abs*{y} \leq \Vmax\) and \(f(s, a) \leq \Vmax\) for all \(s, a\), (c) our construction of \(\cD\) implies that for each iteration $t$, the sample $(x_t,y_t)$ are generated in the following procedure: $x_t$ is sampled from the data generation scheme $\cD^t(x_{1:t-1},y_{1:t-1})$, and $y_t$ is sampled from some conditional probability distribution $p(\cdot \mid x_t)$ as defined in \pref{lem:sq_loss_generalization}, finally (d) the  samples in \(\cD^\nu_h\) are drawn from the offline distribution \(\nu_h\), and the samples in \(\cD^\tau_{h}\) are drawn such that \(s_h \sim d_h^{\pi^t}\) and \(a_h \sim \pi^{f^t}(s_h)\).
Thus, using \pref{lem:sq_loss_generalization}, we get that the least square regression solution \(f^{t+1}_h\) satisfies 
\begin{align*}
 \sum_{i=1}^n \En \brk*{\prn{f^{t+1}_{h}(s^i, a^i) - \dynamicsopt f^{t+1}_{h+1}(s^i, a^i)}^2 \mid \cD_{i}} &\leq 256 \Vmax^2 \log(2 \abs{\cF}/\delta). 
\end{align*}
Using the property-(d) in the above, we get that
\begin{align*}
\moff \cdot \nrm*{f_h^{t+1} - \dynamicsopt f_{h+1}^{t+1}}_{2, \nu_h}^2 + \mon \cdot  \sum_{\tau = 1}^t  \nrm*{f_h^{t+1} - \dynamicsopt f_{h+1}^{t+1}}^2_{2, \mu^\tau_h} \leq 256 \Vmax^2 \log(2 \abs{\cF}/\delta), 
\end{align*} where the distribution \(\mu^\tau_h \in \Delta(s, a)\) is defined by sampling \(s \sim d^{\pi^\tau}_h\) and \(a \sim \pi^{f^t}(s)\). Taking a union bound over \(h \in [H-1]\) and \(t \in [T]\), and bounding each term separately, gives the desired statement. 
\end{proof}

We next note a change in distribution lemma which allows us to bound expected bellman error under the $(s,a)$ distribution generated by \(f^t\) in terms of the expected square bellman error w.r.t. the previous policies data distribution, which is further controlled using regression. 
\begin{lemma}
\label{lem:online_shift}
For any \(t \geq 0\) and \(h \in [H-1]\), we have 
\begin{align*}
\abs*{\tri*{ W_h(f^t) , X_h(f^t) }} \leq \|X_h(f^t)\|_{\Sigma_{t-1; h}^{-1}} \sqrt{\sum_{i = 1}^{t-1} \En_{s, a \sim d^{f^i}_h} \brk*{ \prn*{ f^t_h - \dynamicsopt f^t_{h+1}}^2} + \lambda B^2_W}, 
\end{align*} 
where $\Sigma_{t - 1}^{-1}$ is defined in \pref{eq:sigma_defn} and use the notation \(d^{f^i}_h\) to denote \(d^{\pi^{f^i}}_h\).
\end{lemma} 
\begin{proof}
Using Cauchy-Schwarz inequality, we get that 
\begin{align*}
\abs*{\tri*{ W_h(f^t) , X_h(f^t) }} &\leq  \|X_h(f^t)\|_{\Sigma_{t-1; h}^{-1}}\|W_h(f^t) \|_{\Sigma_{t-1;h}} \\ 
&= \|X_h(f^t)\|_{\Sigma_{t-1; h}^{-1}} \sqrt{\prn*{W_h(f^t)}^\top \Sigma_{t-1} W_h(f^t)} \\ 
&= \|X_h(f^t)\|_{\Sigma_{t-1; h}^{-1}} \sqrt{\prn*{W_h(f^t)}^\top \prn*{ \sum_{i = 1}^{t-1} X_h(f^i) X_h(f^i)^\top + \lambda \bbI } W_h(f^t)}  \\ 
&= \|X_h(f^t)\|_{\Sigma_{t-1; h}^{-1}} \sqrt{\sum_{i = 1}^{t-1}  \abs*{\tri*{W_h(f^t), X_h(f^i)}}^2 + \lambda   \nrm{W_h(f^t)}^2}  \\ 
&\leq  \|X_h(f^t)\|_{\Sigma_{t-1; h}^{-1}} \sqrt{\sum_{i = 1}^{t-1}  \abs*{\tri*{W_h(f^t), X_h(f^i)}}^2 + \lambda   B_W^2} \numberthis \label{eq:part_eqn1} \\ 
&\leq \|X_h(f^t)\|_{\Sigma_{t-1; h}^{-1}} \sqrt{\sum_{i = 1}^{t-1} \En_{s, a \sim d^{f^i}_h} \brk*{ \prn*{ f^t_h - \dynamicsopt f^t_{h+1}}^2} + \lambda B_W^2} 
\end{align*} where the inequality in the second last line holds by plugging in the bound on \(\nrm{W_h(f^t)}\), and the last line holds by using \pref{def:bilinear_model} which implies that 
\begin{align*} 
 \abs*{\tri*{W_h(f^t), X_h(f^i)}}^2 &=  \prn*{\En_{s, a \sim d^{f^i}_h} \brk*{f^t_h - \dynamicsopt f^t_{h+1}}}^2 \leq \En_{s, a \sim d^{f^i}_h}  \brk*{\prn*{f^t_h - \dynamicsopt f^t_{h+1}}^2}, 
\end{align*} where the last inequality is due to Jensen's inequality.  
\end{proof} 

We now have all the tools to prove \pref{thm:main_regret}. We first restate the bound with the exact problem dependent parameters, assumign that \(B_W\) and \(B_X\) are constants which are hidden in the order notation below. 
\begin{theorem*}[\pref{thm:main_regret} restated] Let \(\moff = T\) and \(\mon =1\). Then, with probability at least \(1- \delta\), the cumulative suboptimality of \pref{alg:fqi} is bounded as 
\begin{align*}
\sum_{t = 1}^T V^{\pi^e} - V^{\pi^{f^t}} = O\prn*{ \max\crl*{C_{\pi^e}, 1} \Vmax \sqrt{d H^2 T \cdot \log\prn*{1 + \frac{T}{d}} \log\prn*{\frac{H T \abs{\cF}}{\delta}}}}.
\end{align*}
\end{theorem*} 
\begin{proof}[Proof of \pref{thm:main_regret}] Let \(\pi^e\) be any comparator policy with bounded transfer coefficient i.e. 
\begin{align*} 
C_{\pi^e} := \max\crl*{0, ~ \max_{f\in \Fcal} \frac{\sum_{h=0}^{H-1} \EE_{s,a\sim d_h^{\pi^e} }  \left[ f_h(s,a) - \dynamicsopt f_{h+1}(s,a) \right] }{ \sqrt{ \sum_{h=0}^{H-1} \EE_{s,a\sim \nu_h }  \brk*{\left( f_h(s,a) - \dynamicsopt f_{h+1}(s,a) \right)^2} }}} < \infty.  \numberthis \label{eq:transfer coefficient_definition}
\end{align*}

We start by noting that  
\begin{align*}
    \sum_{t = 1}^T V^{\pi^e} - V^{\pi^{f^t}} &= \sum_{t = 1}^T  \EE_{s \sim d_0}\brk*{V^{\pi^e}_0(s) - V^{\pi^{f^t}}_0(s)} \\ 
    &= \sum_{t = 1}^T  \EE_{s \sim d_0}\brk*{V^{\pi^e}_0(s) - \max_a f_0^t(s, a)} + \sum_{t = 1}^T  \EE_{s \sim d_0}\brk*{ \max_a f_0^t(s, a) - V^{\pi^{f^t}}_0(s)}. \numberthis \label{eq:main_proof1} 
\end{align*} 

For the first term in the right hand side of \pref{eq:main_proof1}, note that using \pref{lem:optimism} for each \(f_t\)  for \(1 \leq t \leq T \), we get   
\begin{align*}
\sum_{t=1}^T \EE_{s \sim d_0}  \brk*{ V^{\pi^e}_0(s) - \max_a f^t_0(s, a)} &\leq \sum_{t=1}^T \sum_{h= 0}^{H - 1}\EE_{s,a \sim d^{\pi_e}_h}[\dynamicsopt  f^t_{h+1}(s,a) - f^t_h(s,a)] \\
&\leq \sum_{t=1}^T C_{\pi^e} \cdot  \sqrt{\sum_{h=0}^{H-1} \En_{s, a \sim \nu_h} \brk*{\prn*{f_h^t(s, a) - \dynamicsopt f_{h + 1}^t(s, a)}^2}} \\
&= T C_{\pi^e} \cdot  \sqrt{H \cdot \Delta_{\mathrm{off}}}, \numberthis \label{eq:regret1}
\end{align*} 
where the second inequality follows from plugging in the definition of \(C_{\pi_e}\) in   \pref{eq:transfer coefficient_definition}. The last line follows from  \pref{lem:fqi_guarantee}. 

For the second term in \pref{eq:main_proof1}, using \pref{lem:simulation} for each \(f_t\) for \(1 \leq t \leq T\), we get  
\begin{align*}
\sum_{t = 1}^T  \EE_{s \sim d_0}\brk*{ \max_a f_0^t(s, a) - V^{\pi^{f^t}}_0(s)} &\leq \sum_{t=1}^T \sum_{h = 0}^{H - 1} \abs*{ \EE_{s,a \sim d_h^{\pi^{f^t}}}\brk*{f^t_h(s,a) - \Tcal f^t_{h+1}(s,a)}} \numberthis \label{eq:checkpoint_1} \\
&= \sum_{t=1}^T \sum_{h=0}^{H-1} \abs*{\tri*{X_h(f^t), W_h(f^t)}} \\  
&\leq \sum_{t=1}^T \sum_{h=0}^{H-1} \|X_h(f^t)\|_{\Sigma_{t-1; h}^{-1}} \sqrt{\Delta_{\mathrm{on}} + \lambda B_W^2},  
\end{align*}
where the second line follows from \pref{def:bilinear_model}, the third line follows from \pref{lem:online_shift} and by plugging in the bound in \pref{lem:fqi_guarantee}. Using the bound in \pref{lem:elliptical} in the above, we get that 
\begin{align*}
\sum_{t = 1}^T  \EE_{s \sim d_0}\brk*{ \max_a f_0^t(s, a) - V^{\pi^{f^t}}_0(s)} &\leq \sqrt{2 d H ^2 \log\prn*{1 + \frac{TB_X^2}{\lambda d}} \cdot \prn*{\Delta_{\mathrm{on}} + \lambda B_W^2} \cdot T} \\
&\leq \sqrt{2 d H ^2 \log\prn*{1 + \frac{T}{d}} \cdot \prn*{\Delta_{\mathrm{on}} + B_X^2 B_W^2} \cdot T} 
,  \numberthis \label{eq:regret2}
\end{align*} where the second line follows by plugging in \(\lambda = B_X^2\). 

Combining the bound \pref{eq:regret1} and \pref{eq:regret2}, we get that 
\begin{align*}
    \sum_{t = 1}^T V^{\pi^e} - V^{\pi^{f^t}} &\leq T C_{\pi^e} \cdot  \sqrt{H \cdot \Delta_{\mathrm{off}}} + \sqrt{2 d H ^2 \log\prn*{1 + \frac{T}{d}} \cdot \prn*{\Delta_{\mathrm{on}} + B_X^2 B_W^2} \cdot T} 
\end{align*} 

Plugging in the values of \(\Delta_{\mathrm{on}}\) and \(\Delta_{\mathrm{off}}\) in the above, and using subadditivity of square-root, we get that 
\begin{align*}
    \sum_{t = 1}^T V^{\pi^e} - V^{\pi^{f^t}} &\leq 16\Vmax C_{\pi^e} T   \sqrt{\frac{H}{\moff} \log\prn*{\frac{2 H T \abs{\cF}}{\delta}}} + 16 \Vmax \sqrt{\frac{2 dH^2 T}{\mon} \log\prn*{1 + \frac{T}{d}} \log\prn*{\frac{2 H T \abs{\cF}}{\delta}}} \\
    & \qquad \qquad \qquad + H B_X B_W \sqrt{2d T \log\prn*{1 + \frac{T}{d}}}. 
\end{align*}

Setting \(\moff = T\) and \(\mon = 1\) in the above gives the cumulative suboptimality bound 
\begin{align*}
\sum_{t = 1}^T V^{\pi^e} - V^{\pi^{f^t}} = O\prn*{ \max\crl*{C_{\pi^e}, 1} \Vmax \sqrt{d H^2 T \cdot \log\prn*{1 + \frac{T}{d}} \log\prn*{\frac{H T \abs{\cF}}{\delta}}}}. \numberthis \label{eq:final_bound} 
\end{align*} 
\end{proof}

\begin{proof}[Proof of \pref{corr:sample_complexity}]  We next convert the above cumulative suboptimality bound into sample complexity bound via a standard online-to-batch conversion. Setting \(\pi^e = \pi^*\) in \pref{eq:final_bound} and defining the policy \(\wh \pi = \text{Uniform}\prn*{\crl*{\pi^1, \dots, \pi^T}}\), we get that 
\begin{align*} 
\En \brk*{V^{\pi^*} - V^{\wh \pi}} &= \frac{1}{T} \prn*{\sum_{t = 1}^T V^{\pi^*} - V^{\pi^{t}}} \\
&= O\prn*{ \max\crl*{C_{\pi^*}, 1} \Vmax \sqrt{\frac{d H^2}{T} \cdot \log\prn*{1 + \frac{T}{d}} \log\prn*{\frac{H T \abs{\cF}}{\delta}}}}. 
\end{align*} 

Thus, we get that for \(T \geq \wt{O} \prn*{ \frac{\max\crl*{C^2_{\pi^*}, 1} \Vmax^2 d H^2 \log\prn*{{H T \abs{\cF}}/{\delta}}}{\epsilon^2}}\), we get that $$\En \brk*{V^{\pi^*} - V^{\wh \pi}} \leq \epsilon.$$ In these \(T\) iterations, the total number of offline samples used is
\begin{align*}
\moff = T = \wt{O} \prn*{ \frac{ \max\crl*{C^2_{\pi^*}, 1} \Vmax^2 d H^2 \log\prn*{{H T \abs{\cF}}/{\delta}}}{\epsilon^2}},  
\end{align*}
and the total number of online samples used is 
\begin{align*}
\mon \cdot H \cdot T = \wt{O} \prn*{ \frac{\max\crl*{C^2_{\pi^*}, 1} \Vmax^2 d H^3 \log\prn*{{H T \abs{\cF}}/{\delta}}}{\epsilon^2}}, 
\end{align*}
where the additional \(H\) factor appears because we collect \(\mon\) samples for every \(h \in [H]\) in the algorithm.
\end{proof} 

\subsection{V-type Bilinear Rank} \label{app:V_type_appendix}
Our previous result focus on the Q-type bilinear model.
Here we provide the V-type Bilinear rank definition. This V-type Bilinear rank definition is basically the same as the low Bellman rank model proposed by \cite{jiang2017contextual}.
\begin{definition}[{V-type Bilinear model}] 
\label{def:V_type_bilinear_model}
Consider any pair of functions $(f, g)$ with $f,g\in \Fcal$. Denote the greedy policy of $f$ as $\pi^f = \{ \pi^f_h := \argmax_{a} f_h(s,a), \forall h\}$. We say that the MDP together with the function $\Fcal$ admits a bilinear structure of rank \(d\) if for any \(h \in [H-1]\), there exist two (unknown) mappings $X_h:\Fcal\mapsto \mathbb{R}^d$ and $W_h:\Fcal\mapsto \mathbb{R}^d$ with $\max_{f}\| X_h(f)\|_2 \leq B_X$ and $\max_{f} \|W_h(f)\|_2 \leq B_W$, such that:
\begin{align*}
\forall f, g\in \Fcal: \; \abs*{\EE_{s\sim d_h^{\pi^f}, a\sim \pi_{g}(s)}  g_h(s,a) - \dynamicsopt g_{h+1}(s,a) } = \left\lvert \tri*{ X_h(f), W_h(g) }  \right\rvert.
\end{align*} 
\end{definition}
Note that different from the Q-type definition, here the action $a$ is taken from the greedy policy with respect to $g$. This way $\max_{a} g(s,a)$ can serve as an approximation of $V^\star$ -- thus the name of $V$-type. 

To make \algname{} work for the V-type Bilinear model, we only need to make slight change on the data collection process, i.e., when we collect online batch $\Dcal_h$, we sample $s\sim d^{\pi^t}_h, a\sim \text{Uniform}(\Acal), s'\sim P(\cdot | s,a)$. Namely the action is taken uniformly randomly here. We provide the pseudocode in \pref{alg:vtype}. We refer the reader to \cite{du2021bilinear,jin2021bellman} for a detailed discussion. 

\begin{algorithm}[t] 
\caption{V-type Hy-Q}
\begin{algorithmic}[1] 
\REQUIRE Value function class: $\Fcal$, \#iterations: $T$, Offline dataset \(\cD^{\nu}_h\) of size \(m_\mathrm{off}\) for \(h \in [H-1]\). 
\\
\STATE Initialize \(f_h^1(s, a) = 0\).
\FOR{$t = 1, \dots, T$}  
\STATE Let $\pi^t$ be the greedy policy w.r.t. \(f^t\) i.e., $\pi_h^t(s) = \argmax_a f^t_h(s,a).$ \\
 \vspace{1mm} 
\algcomment{Online collection}\\
 \vspace{1mm} 
\STATE For each $h$, collect $m_{\mathrm{on}}$ online tuples $\cD^t_h \sim d_h^{\pi^t}\circ \textrm{Uniform}(\cA)$. \label{line:onlinev} \\
 \vspace{1mm} 
 \algcomment{FQI using both online and offline data}
 \vspace{1mm}
 \label{line:online_sample}
\STATE Set $f_H^{t+1}(s,a) = 0$.  \\ 
\FOR{$h = H-1, \dots, 0$}  
\STATE Estimate \(f_h^{t+1}\) using least squares regression on the aggregated data: 
{\small \begin{align*}
\hspace{-10mm} f_{h}^{t+1} \leftarrow \argmin_{f\in \Fcal_h}   \crl*{ 
 \wh \EE_{\Dcal^\nu_{h}}( f(s,a) -   r -  \max_{a'} f^{t+1}_{h+1}(s', a') )^2   +  \sum_{\tau = 1}^{t} \wh \EE_{\Dcal^\tau_{h}}( f(s,a) -   r -  \max_{a'} f^{t+1}_{h+1}(s', a') )^2} \numberthis \label{eq:alg_regression_vtype}
\end{align*}} 
\ENDFOR
\ENDFOR
\end{algorithmic}\label{alg:vtype} 
\end{algorithm} 

\subsubsection{Complexity bound for V-type Bilinear models}
In this section, we give a performance analysis of \pref{alg:vtype} for V-type Bilinear models. The contents in this section extend the results developed for Q-type Bilinear models in \pref{sec:main_proofs} to V-type Bilinear models.

We first note the following bound for FQI estimates in \pref{alg:vtype}. 
\begin{lemma}
\label{lem:V_type_fqi_guarantee}
Let \(\delta \in (0,1)\) and let for \(h \in [H-1]\) and \(t \in [T]\), \(f^{t+1}_h\) be the estimated value function for time step \(h\) computed via least square regression using samples in the dataset \({\prn*{\cD^\nu_h, \cD^{1}_h, \dots, \cD^{t}_h}}\) in \pref{eq:alg_regression_vtype} in the iteration \(t\) of \pref{alg:vtype}. Then, with probability at least \(1 - \delta\), for any \(h \in [H-1]\) and \(t \in [T]\), 
\begin{align*}
\nrm*{f_h^{t+1} - \dynamicsopt f_{h+1}^{t+1}}_{2, \nu_h}^2 \leq \frac{1}{\moff} 256 \Vmax^2 \log(2 H T \abs{\cF}/\delta)  \rdef{} \bar{\Delta}_{\mathrm{off}},
\intertext{and}
\sum_{\tau = 1}^t  \nrm*{f_h^{t+1} - \dynamicsopt f_{h+1}^{t+1}}^2_{2, \mu^\tau_h} \leq \frac{1}{\mon}  256 \Vmax^2 \log(2 H T \abs{\cF}/\delta)  \rdef{} \bar{\Delta}_{\mathrm{on}},
\end{align*} where \(\nu_h\) denotes the offline data distribution at time \(h\), and the distribution \(\mu^\tau_h \in \Delta(s, a)\) is defined such that \(s \sim d^{\pi^\tau}_h\) and \(a \sim \mathrm{Uniform}(\cA)\). 
\end{lemma}

The following change in distribution lemma is the version of \pref{lem:online_shift} under V-type Bellman rank assumption. 
\begin{lemma}
\label{lem:V_type_online_shift}
Suppose the underlying model is a V-type bilinear model. Then, for any \(t \geq 0\) and \(h \in [H-1]\), we have 
\begin{align*}
\abs*{\tri*{ W_h(f^t) , X_h(f^t) }} \leq \|X_h(f^t)\|_{\Sigma_{t-1; h}^{-1}} \sqrt{ \abs*{\cA} \cdot \sum_{i = 1}^{t-1}   \En_{s \sim d_h^{\pi^{f^i}}, ~ a \sim \mathrm{Uniform}(\cA)} \brk*{\prn*{f_h^t - \dynamicsopt f_{h+1}^t}^2} + \lambda   B_W^2},  
\end{align*} 
where $\Sigma_{t - 1}^{-1}$ is defined in \pref{eq:sigma_defn}. 
\end{lemma} 
\begin{proof} The proof is identical to the proof of \pref{lem:online_shift}. Repeating the analysis till \pref{eq:part_eqn1}, we get that 
\begin{align*}
\abs*{\tri*{ W_h(f^t) , X_h(f^t) }}
&\leq \|X_h(f^t)\|_{\Sigma_{t-1; h}^{-1}} \sqrt{\sum_{i = 1}^{t-1}  \abs*{\tri*{W_h(f^t), X_h(f^i)}}^2 + \lambda B_W^2} \\
&= \|X_h(f^t)\|_{\Sigma_{t-1; h}^{-1}} \sqrt{\sum_{i = 1}^{t-1}  \prn*{\En_{s \sim d_h^{\pi^{f^i}}, ~ a \sim \pi^{f^t}(s)} \brk*{f_h^t - \dynamicsopt f_{h+1}^t}}^2 + \lambda   B_W^2}   \\
&\leq \|X_h(f^t)\|_{\Sigma_{t-1; h}^{-1}} \sqrt{ \abs*{\cA} \cdot \sum_{i = 1}^{t-1}   \En_{s \sim d_h^{\pi^{f^i}}, ~ a \sim \mathrm{Uniform}(\cA)} \brk*{\prn*{f_h^t - \dynamicsopt f_{h+1}^t}^2} + \lambda   B_W^2}   \\
\end{align*} 
where the second line above follows from the definition of V-type bilinear model in \pref{def:V_type_bilinear_model}, and the last line holds because:   
\begin{align*}
\prn*{\En_{s \sim d_h^{\pi^{f^i}}, ~ a \sim \pi^{f^t}(s)} \brk*{f_h^t - \dynamicsopt f_{h+1}^t}}^2 &\leq   \En_{s \sim d_h^{\pi^{f^i}},  a \sim \pi^{f^t}(s)} \brk*{\prn*{f_h^t - \dynamicsopt f_{h+1}^t}^2} \\
&\leq \abs*{\cA} \cdot \En_{s \sim d_h^{\pi^{f^i}},  a \sim \mathrm{Uniform}(\cA)} \brk*{\prn*{f_h^t - \dynamicsopt f_{h+1}^t}^2}
\end{align*} where the first inequality above is due to Jensen's inequality and the last  inequality follows form a straightforward upper bound since each term inside the expectation is non-negative. 
\end{proof} 

We are finally ready to state and prove our main result in this section. 

\begin{theorem}[Cumulative suboptimality bound for V-type bilinear rank models] 
\label{thm:Vtype_total_complexity}
Let  \(\mon = \abs*{\cA}\) and \(\moff = T\). Then, with probability at least \(1- \delta\), the cumulative suboptimality of \pref{alg:vtype} is bounded as 
\begin{align*}
\sum_{t = 1}^T V^{\pi^e} - V^{\pi^{f^t}} = O\prn*{ \max\crl*{C_{\pi^e}, 1} \Vmax \sqrt{d H^2 T \cdot \log\prn*{1 + \frac{T}{d}} \log\prn*{\frac{H T \abs{\cF}}{\delta}}}}
\end{align*}
\end{theorem} 
\begin{proof} The proof follows closely the proof of \pref{thm:main_regret}. Repeating the analysis till \pref{eq:main_proof1}  and \pref{eq:regret1}, we get that: 
\begin{align*}
    \sum_{t = 1}^T V^{\pi^e} - V^{\pi^{f^t}} 
    & \leq T C_{\pi^e} \cdot  \sqrt{H \cdot \bar{\Delta}_{\mathrm{off}}} + \sum_{t = 1}^T  \EE_{s \sim d_0}\brk*{ \max_a f_0^t(s, a) - V^{\pi^{f^t}}_0(s)}. \numberthis \label{eq:Vtype_main_proof1} 
\end{align*} 

For the second term in the above, using \pref{lem:simulation} for each \(f_t\) for \(1 \leq t \leq T\), we get  
\begin{align*}
\sum_{t = 1}^T  \EE_{s \sim d_0}\brk*{ \max_a f_0^t(s, a) - V^{\pi^{f^t}}_0(s)} &\leq \sum_{t=1}^T \sum_{h = 0}^{H - 1} \abs*{ \EE_{s,a \sim d_h^{\pi^{f^t}}}\brk*{f^t_h(s,a) - \Tcal_{h}f^t_{h+1}(s,a)}} \\
&= \sum_{t=1}^T \sum_{h=0}^{H-1} \abs*{\tri*{X_h(f^t), W_h(f^t)}} \\  
&\leq \sum_{t=1}^T \sum_{h=0}^{H-1} \|X_h(f^t)\|_{\Sigma_{t-1; h}^{-1}} \sqrt{\abs*{\cA} \cdot \bar{\Delta}_{\mathrm{on}} + \lambda B_W^2},  
\end{align*}
where the second line follows from \pref{def:V_type_bilinear_model}, and the last line follows from \pref{lem:V_type_online_shift} and by plugging in the bound in  \pref{lem:V_type_fqi_guarantee}. Using the elliptical potential \pref{lem:elliptical} as in the proof of \pref{thm:main_regret}, we get that 
\begin{align*}
    \sum_{t = 1}^T V^{\pi^e} - V^{\pi^{f^t}} &\leq T C_{\pi^e} \cdot  \sqrt{H \cdot \bar{\Delta}_{\mathrm{off}}} + \sqrt{2 d H ^2 \log\prn*{1 + \frac{T}{d}} \cdot \prn*{\abs*{\cA} \cdot \bar{\Delta}_{\mathrm{on}} + B_X^2 B_W^2} \cdot T} 
\end{align*}

Plugging in the values of \(\bar{\Delta}_{\mathrm{on}}\) and \(\bar{\Delta}_{\mathrm{off}}\) from \pref{lem:V_type_fqi_guarantee} in the above, and using subadditivity of square-root, we get that 
\begin{align*}
    \sum_{t = 1}^T V^{\pi^e} - V^{\pi^{f^t}} 
    &\leq 16\Vmax C_{\pi^e} T   \sqrt{\frac{H}{\moff} \log\prn*{\frac{2 H T \abs{\cF}}{\delta}}} + 16 \Vmax \sqrt{\frac{2 dH^2 \abs*{\cA} T}{\mon} \log\prn*{1 + \frac{T}{d}} \log\prn*{\frac{2 H T \abs{\cF}}{\delta}}} \\
    & \qquad \qquad \qquad + H B_X B_W \sqrt{2d T \log\prn*{1 + \frac{T}{d}}}. 
\end{align*}

Setting \(\mon = \abs*{\cA}\) and \(\moff = T\), we get the following cumulative suboptimality bound: 
\begin{align*}
\sum_{t = 1}^T V^{\pi^e} - V^{\pi^{f^t}} = O\prn*{ \max\crl*{C_{\pi^e}, 1} \Vmax \sqrt{d H^2 T \cdot \log\prn*{1 + \frac{T}{d}} \log\prn*{\frac{H T \abs{\cF}}{\delta}}}}. \numberthis \label{eq:Vtype_final_bound}
\end{align*}
\end{proof}

\begin{corollary}[Sample complexity] 
\label{corr:Vtype_sample_complexity} 
Under the assumptions of \pref{thm:Vtype_total_complexity} if \(C_{\pi^*} < \infty\) then \pref{alg:vtype} can find an \(\epsilon\)-suboptimal policy \(\wh \pi\) for which \(V^{\pi^*} - V^{\wh \pi} \leq \epsilon\) with total sample complexity of:  
\begin{align*} 
n = \wt{O} \prn*{ \frac{\max\crl*{C^2_{\pi^*}, 1} \Vmax^2 d  H^3 \abs*{\cA} \log\prn*{{H T \abs{\cF}}/{\delta}}}{\epsilon^2}}. 
\end{align*}
\end{corollary}
\begin{proof} The following follows from a standard online-to-batch conversion. Setting \(\pi^e = \pi^*\) in \pref{eq:Vtype_final_bound} and defining the policy \(\wh \pi = \text{Uniform}\prn*{\crl*{\pi^1, \dots, \pi^T}}\), we get that 
\begin{align*} 
\En \brk*{V^{\pi^*} - V^{\wh \pi}} &= \frac{1}{T} \prn*{\sum_{t = 1}^T V^{\pi^*} - V^{\pi^{t}}} = O\prn*{ \max\crl*{C_{\pi^e}, 1} \Vmax \sqrt{\frac{d H^2}{T} \cdot \log\prn*{1 + \frac{T}{d}}  \log\prn*{\frac{H T \abs{\cF}}{\delta}}}}. 
\end{align*} 

Thus, we the policy returned after \(T \geq \wt{O} \prn*{ \frac{\max\crl*{C^2_{\pi^*}, 1} \Vmax^2 d H^2 \log\prn*{{H T \abs{\cF}}/{\delta}}}{\epsilon^2}}\) satisfies $\En \brk*{V^{\pi^*} - V^{\wh \pi}} \leq \epsilon.$ In these \(T\) iterations, the total number of offline samples used is
\begin{align*}
\moff = T = \wt{O} \prn*{ \frac{\max\crl*{C^2_{\pi^*}, 1} \Vmax^2 d H^2 \log\prn*{{H T \abs{\cF}}/{\delta}}}{\epsilon^2}}, 
\end{align*}
and the total number of online samples collected is 
\begin{align*}
\mon \cdot H \cdot T = \wt{O} \prn*{ \frac{\max\crl*{C^2_{\pi^*}, 1} \Vmax^2 d  H^3 \abs*{\cA} \log\prn*{{H T \abs{\cF}}/{\delta}}}{\epsilon^2}}, 
\end{align*} 
where the additional \(H\) factor appears because we collect \(\mon\) samples for every \(h \in [H]\) in the algorithm.
\end{proof}

\subsection{Bounds on transfer coefficient}\label{app:coef}
Note that \(C_{\pi}\) takes both the distribution shift and the function class into consideration, and is smaller than the existing density ratio based concentrability coefficient \citep{kakade2002approximately, munos2008finite, chen2019information} and also existing Bellman error based concentrability coefficient  \cite{xie2021bellman}. We formalize this in the following lemma. 

\begin{lemma}\label{lem:coefcon}For any \(\pi\) and offline distribution \(\nu\), 
\begin{align*}
C_{\pi} &\leq \sqrt{\max_{f, h} \frac{\nrm{f_h - \dynamicsopt f_{h+1}}^2_{d_h^\pi}}{\nrm{f_h - \dynamicsopt f_{h+1}}^2_{\nu_h}}} \leq \sup_{h, s,a} \frac{d_h^\pi(s, a)}{\nu_h(s, a)}. 
\end{align*}
\end{lemma}

\begin{proof} Using Jensen's inequality, we get that 
\begin{align*}
C_{\pi} &\leq \sqrt{\max_{f} \frac{\sum_{h=0}^{H-1} \nrm{f_h - \dynamicsopt f_{h+1}}^2_{d_h^\pi}}{\sum_{h=0}^{H-1}  \nrm{f_h - \dynamicsopt f_{h+1}}^2_{\nu_h}}} \\
&\leq  \sqrt{\max_{f, h} \frac{\nrm{f_h - \dynamicsopt f_{h+1}}^2_{d_h^\pi}}{\nrm{f_h - \dynamicsopt f_{h+1}}^2_{\nu_h}}} \\
&\leq \sqrt{ \sup_{h, s,a} \frac{d_h^\pi(s, a)}{\nu_h(s, a)} } \\
&\leq \sup_{h, s,a} \frac{d_h^\pi(s, a)}{\nu_h(s, a)}, 
\end{align*} where the second line follows from the Mediant inequality and the last line holds whenever \(\sup_{h, s,a} \frac{d_h^\pi(s, a)}{\nu_h(s, a)} \geq 1\). 
\end{proof}

Next we show that in the linear Bellman complete setting, $C_{\pi}$ is bounded by the relative condition number using the linear features. 
\begin{lemma}
Consider the linear Bellman complete setting (\pref{def:linear_bellman_completeness}) with known feature $\phi$. Suppose that the feature covariance matrix induced by offline distribution $\nu$: $\Sigma_{\nu_{h}} \ldef{} \EE_{s,a \sim \nu_{h}}[\phi^\star(s,a)\phi^\star(s,a)^\top]$ is invertible. Then for any policy \(\pi\), we have 
\begin{align*}
   C_{\pi} &\leq \sqrt{\max_h \EE_{s,a \sim d^{\pi}_{h}}\nrm{\phi(s,a)}^2_{\Sigma_{\nu_{h}}^{-1}}}.
\end{align*}
\end{lemma}
\begin{proof}
Repeating the argument in \pref{lem:coefcon}, we have
\begin{align*}
C_{\pi} &\leq \sqrt{\max_{f,h} \frac{ \nrm{f_h - \dynamicsopt f_{h+1}}^2_{d_h^\pi}}{ \nrm{f_h - \dynamicsopt f_{h+1}}^2_{\nu_h}}} \\
&\leq \sqrt{\max_{w,h} \frac{ \nrm{w_h^{\top} \phi - {w'}_h^{\top} \phi}^2_{d_h^\pi}}{ \nrm{w_h^{\top} \phi - {w'}_h^{\top} \phi}^2_{\nu_h}}} \\
&\leq \sqrt{\max_{w,h} \frac{ \nrm{(w_h-w'_h)}_{\Sigma_{\nu_h}}^2 \EE_{d^{\pi}_h} \nrm{\phi}^2_{\Sigma_{\nu_h}^{-1}}}{ \nrm{(w_h-w'_h)^{\top} \phi}^2_{\nu_h}}} \\
&= \sqrt{\max_h\EE_{s,a \sim d^{\pi}_{h}}\nrm{\phi(s,a)}^2_{\Sigma_{\nu_{h}}^{-1}}}.
\end{align*}
Recall that in linear Bellman complete setting, we can write $f$ as $w^{\top}\phi$, and for any $w$ that defines $f$, there exists $w'$ such that $\dynamicsopt f = w'^{\top}\phi$. 
\end{proof}

Now we proceed to low-rank MDPs where feature is unknown. We show that for low-rank MDPs, \(C_{\pi}\) is bounded by the partial feature coverage using the unknown ground truth feature. 
\label{app:transfer_coefficient_bound}
\begin{lemma}\label{lem:low_rank_mdp_transfer}Consider the low-rank MDP setting (\pref{def:low_rank_mdp}) where the transition dynamics \(P\) is given by \(P(s' \mid s, a) = \tri*{\mu^\star(s'), \phi^\star(s, a)}\) for some \(\mu^\star, \phi^\star \in \bbR^d\). Suppose that the offline distribution \(\nu = \prn{\nu_0, \dots, \nu_{H-1}}\) is such that $\max_h \max_{s,a} \frac{\pi_{h}(a|s)}{\nu_h(a|s)} \le \alpha$ for any \(s, a\). Furthermore, suppose that \(\nu\) is induced via trajectories i.e. \(\nu_0(s) = d_0(s)\) and $\nu_h(s) = \EE_{\bar{s}, \bar{a} \sim \nu_{h-1}} P(s|\bar{s}, \bar{a})$ for any \(h \geq 1\), and that the feature covariance matrix  $\Sigma_{\nu_{h-1}, \phi^\star} \ldef{} \EE_{s,a \sim \nu_{h-1}}[\phi^\star(s,a)\phi^\star(s,a)^\top]$ is invertible.\footnote{This is for notation simplicity, and we emphasize that we do not assume eigenvalues are lower bounded. In other words, eigenvalue of this feature covariance matrix could approach to $0^+$.} Then for any policy \(\pi\), we have 
\begin{align*}
C_{\pi} &\leq \sqrt{\alpha} \sum_{h=1}^H \EE_{s, a \sim d_{h-1}^{\pi}} \brk*{ \nrm*{\phi^\star(s, a)}_{\Sigma_{\nu_{h-1}, \phi^\star}^{-1}}} + \sqrt{\alpha}. 
\end{align*} 
\end{lemma}
\begin{proof} We first upper bound the numerator separately. First note that for \(h = 0\), 
\begin{align*}
    \EE_{s,a \sim d_0^{\pi} }  \left[\dynamicsopt f_{1}(s,a) - f_0(s,a) \right] &\leq \sqrt{\EE_{s \sim d_0, a \sim \pi(\cdot|s)}  \left[ \prn*{\dynamicsopt f_{1}(s,a) - f_0(s,a)}^2 \right]}\\
    &\leq \sqrt{\max_{s,a} \frac{d^{\pi}_0(s,a)}{\nu_0(s,a)} \cdot \EE_{s,a\sim \nu_0 }  \left[\prn*{\dynamicsopt f_{1}(s,a) - f_0(s,a)}^2 \right]}\\
    &\leq \sqrt{\alpha \cdot \EE_{s,a\sim \nu_0 } \left[\prn*{ \dynamicsopt f_{1}(s,a) - f_0(s,a)}^2 \right]}, \numberthis \label{eq:linear_mdp_0}
\end{align*} where the last inequality follows from our assumption since \(\max_{s,a} \frac{d^{\pi}_0(s,a)}{\nu_0(s,a)}  = \max_{s,a} \frac{\pi_0(a|s)}{\nu_0(a|s)} \leq \alpha\).  

Next, for any \(h \geq 1\), we note that backing up one step and looking at the pair \(\bar{s}, \bar{a}\) that lead to the state \(s\), we get that  
\begin{align*}
\hspace{0.4in} & \hspace{-0.3in} \EE_{s,a\sim d_h^{\pi} }  \left[  \dynamicsopt f_{h+1}(s,a) - f_h(s,a) \right]  \\
    &= \EE_{\bar{s}, \bar{a} \sim d_{h-1}^{\pi}, s \sim \transitionopt(\bar{s}, \bar{a}), a \sim \pi(s)}  \left[  \dynamicsopt f_{h+1}(s,a) - f_h(s,a) \right] \\
    &= \EE_{\bar{s}, \bar{a} \sim d_{h-1}^{\pi}} \brk*{\int \prn*{\phi^\star(\bar{s}, \bar{a})^\top \mu^\star(s)} \sum_a  \pi(a|s)\left[  \dynamicsopt f_{h+1}(s,a) - f_h(s,a) \right]\text{d}s} \\
    &= \EE_{\bar{s}, \bar{a} \sim d_{h-1}^{\pi}} \brk*{\phi^\star(\bar{s}, \bar{a})^\top \int \sum_a \mu^\star(s)\pi(a|s)\left[  \dynamicsopt f_{h+1}(s,a) - f_h(s,a) \right]\text{d}s} \\
    &\leq \EE_{\bar{s}, \bar{a} \sim d_{h-1}^{\pi}} \brk*{ \nrm*{\phi^\star(\bar{s}, \bar{a})}_{\Sigma_{\nu_{h-1}, \phi^\star}^{-1}} \nrm*{\int \sum_a \mu^\star(s)\pi(a|s)\left[  \dynamicsopt f_{h+1}(s,a) - f_h(s,a) \right]\mathrm{d}s}_{\Sigma_{\nu_{h-1}, \phi^\star}} }, \numberthis \label{eq:linear_mdp_1}
\end{align*} 
where the last line follows from an application of Cauchy-Schwarz inequality. For the term inside the expectation in the right hand side above, we note that, 
\begin{align*}
    \hspace{0.5in}& \hspace{-0.5in} \nrm*{\int \sum_a \mu^\star(s)\pi(a|s)\left[  \dynamicsopt f_{h+1}(s,a) - f_h(s,a) \right]\mathrm{d}s}^2_{\Sigma_{\nu_{h-1}, \phi^\star}}\\
    &\overeq{\proman{1}} \En_{\bar{s}, \bar{a} \sim \nu_{h-1}} \brk*{\prn*{\int \sum_a \prn*{\mu^\star(s)^\top \phi^*(\bar{s}, \bar{a})} \pi(a|s)\prn*{  \dynamicsopt f_{h+1}(s,a) - f_h(s,a)} \mathrm{d}s}^2} \\
    &= \En_{\bar{s}, \bar{a} \sim \nu_{h-1}} \brk*{\prn*{\En_{s \sim P(\bar{s}, \bar{a}), a \sim \pi(s)} \brk*{ \dynamicsopt f_{h+1}(s,a) - f_h(s,a)}}^2} \\
    &\overleq{\proman{2}} \En_{\bar{s}, \bar{a} \sim \nu_{h-1}, s \sim P(\bar{s}, \bar{a}), a \sim \pi(s)} \brk*{ \prn*{ \dynamicsopt f_{h+1}(s,a) - f_h(s,a)}^2}  \\ 
    &\overeq{\proman{3}} \En_{s \sim \nu_{h}, a \sim \pi(s)}   \brk*{ \prn*{ \dynamicsopt f_{h+1}(s,a) - f_h(s,a)}^2}  \\
    &\overleq{\proman{4}} \alpha \cdot \En_{s, a \sim \nu_{h}}  \brk*{ \prn*{ \dynamicsopt f_{h+1}(s,a) - f_h(s,a)}^2}  \numberthis \label{eq:linear_mdp_2}
    \end{align*} where \(\proman{1}\) follows by expanding the norm  
    , $\proman{2}$ follows an application of Jensen's inequality, \(\proman{3}\) is due to our assumption that the offline dataset is generated using trajectories such that \(\nu_h(s) = \En_{\bar{s}, \bar{s} \sim \nu_{h-1}} \brk*{P(s \mid \bar{s}, \bar{a})}\). Finally, \(\proman{4}\) follows from the definition of \(\alpha\). Plugging \pref{eq:linear_mdp_2} in \pref{eq:linear_mdp_1}, we get that for \(h \geq 1\), 
\begin{align*}
\hspace{0.5in} & \hspace{-0.3in} \EE_{s,a\sim d_h^{\pi} }  \left[  \dynamicsopt f_{h+1}(s,a) - f_h(s,a) \right]  \\
&\leq \EE_{\bar{s}, \bar{a} \sim d_{h-1}^{\pi}} \brk*{ \nrm*{\phi^\star(\bar{s}, \bar{a})}_{\Sigma_{\nu_{h-1}, \phi^\star}^{-1}}  \sqrt{\alpha \cdot \En_{s, a \sim \nu_{h}}  \brk*{ \prn*{ \dynamicsopt f_{h+1}(s,a) - f_h(s,a)}^2} }} \numberthis \label{eq:linear_mdp_3}  
\end{align*}  

We are now ready to bound the transfer coefficient. First note that using \pref{eq:linear_mdp_0}, for any \(f\), 
\begin{align*}
\frac{\EE_{s,a\sim d_0^{\pi} }  \left[ \dynamicsopt f_{1}(s,a) - f_0(s,a) \right] }{ \sqrt{ \sum_{h=0}^{H-1} \EE_{s,a\sim \nu_h }  \brk*{\left(  \dynamicsopt f_{h+1}(s,a) - f_h(s,a) \right)^2} }} &\leq \frac{\sqrt{\alpha \cdot \EE_{s,a\sim \nu_0}  \left[ \prn*{\dynamicsopt f_{1}(s,a) - f_0(s,a)}^2 \right] }}{ \sqrt{ \sum_{h=0}^{H-1} \EE_{s,a\sim \nu_h }  \brk*{\left(  \dynamicsopt f_{h+1}(s,a) - f_h(s,a) \right)^2} }}  \\
&\leq \sqrt{\alpha}. 
\end{align*}

Furthermore, for any \(f\), using \pref{eq:linear_mdp_3}, we get that 
\begin{align*}
\hspace{0.4in} & \hspace{-0.2in}  \frac{\sum_{h=1}^{H-1} \EE_{s,a\sim d_h^{\pi} }  \left[  \dynamicsopt f_{h+1}(s,a) - f_h(s,a) \right] }{ \sqrt{ \sum_{h=0}^{H-1} \EE_{s,a\sim \nu_h }  \brk*{\left(  \dynamicsopt f_{h+1}(s,a) - f_h(s,a) \right)^2} }} \\
&\leq \sum_{h=1}^{H-1} \EE_{\bar{s}, \bar{a} \sim d_{h-1}^{\pi}} \brk*{ \nrm*{\phi^\star(\bar{s}, \bar{a})}_{\Sigma_{\nu_{h-1}, \phi^\star}^{-1}} \frac{ \sqrt{\alpha \cdot \En_{s, a \sim \nu_{h}}  \brk*{ \prn*{ \dynamicsopt f_{h+1}(s,a) - f_h(s,a)}^2} }}{\sqrt{ \sum_{h=0}^{H-1} \EE_{s,a\sim \nu_h }  \brk*{\left(  \dynamicsopt f_{h+1}(s,a) - f_h(s,a) \right)^2} }}} \\
&\leq \sum_{h=1}^H \EE_{\bar{s}, \bar{a} \sim d_{h-1}^{\pi}} \brk*{ \nrm*{\phi^\star(\bar{s}, \bar{a})}_{\Sigma_{\nu_{h-1}, \phi^\star}^{-1}} \sqrt{\alpha}}, 
\end{align*} where the last line holds for an appropriate choice of \(\lambda\) (e.g. \(\lambda = 0\)). Combining the above two bounds in the definition of \(C_{\pi}\) we get that 
\begin{align*}
C_{\pi} &\leq \sqrt{\alpha} \sum_{h=1}^H \EE_{\bar{s}, \bar{a} \sim d_{h-1}^{\pi}} \brk*{ \nrm*{\phi^\star(\bar{s}, \bar{a})}_{\Sigma_{\nu_{h-1}, \phi^\star}^{-1}}} + \sqrt{\alpha}. 
\end{align*} 
\end{proof} 
Note that in the above result, the transfer coefficient is upper bounded by the relative coverage under unknown feature $\phi^\star$ and a term $\alpha$ related to the action coverage, i.e., $\max_h \max_{s,a} \frac{\pi_h(a|s)}{\nu_h(a|s)} \le \alpha$. This matches to the coverage condition used in prior offline RL works for low-rank MDPs \citep{uehara2021pessimistic}.

\section{Auxiliary Lemmas}\label{app:support}
In this section, we provide a few results and their proofs that we used in the previous sections. We first with the following form of Freedman's inequality that is a modification of a similar inequality in \citep{beygelzimer2011contextual}.

\begin{lemma}[Freedman's Inequality] 
\label{lem:freedman_form}
Let \(\crl*{X_1, \dots, X_T}\) be a sequence of non-negative random variables where each $x_t$ is sampled from some process that depends on all previous instances, i.e, $x_t \sim \distrib_t = \distrib_t(x_{1:t-1})$. Further, suppose that \(\abs*{X_t} \leq R\) almost surely for all \(t \leq T\). Then, for any \(\delta > 0\) and \(\lambda \in [0, 1/2R]\), with probability at least \(1 - \delta\), 
\begin{align*}
\abs*{\sum_{t=1}^T X_t -\En\brk*{X_t \mid \distrib_t}} &\leq \lambda \sum_{t=1}^T \prn*{2  R \abs*{\En\brk*{X_t \mid \distrib_t}}  + \En\brk*{X_t^2 \mid \distrib_t}} + \frac{\log(2/\delta)}{\lambda} .
\end{align*}
\end{lemma}
\begin{proof} Define the random variable \(Z_t = X_t - \En\brk*{X_t \mid \distrib_{t}}\). Clearly, \(\crl*{Z_t}_{t=1}^T\) is a martingale difference sequence. 
Furthermore, we have that for any \(t\),  \(\abs*{Z_t} \leq 2R\) and that
\begin{align*}
\En \brk*{Z_t^2 \mid \distrib_t} &=   \En \brk*{\prn*{X_t - \En \brk*{X_t\mid \distrib_t}}^2\mid \distrib_t} \leq 2 R \abs*{\En \brk*{X_t \mid \distrib_t}} + \En \brk*{X_t^2 \mid \distrib_t}.  \numberthis \label{eq:freedman1}
\end{align*} where the last inequality holds because \(\abs*{X_t} \leq R\). 

Using the form of Freedman's inequality in \citet[Theorem 1]{beygelzimer2011contextual}, we get that for any \(\lambda \in [0, 1/2R]\), 
\begin{align*}
\abs*{\sum_{t=1}^T Z_t} \leq \lambda \sum_{t=1}^T \En \brk*{Z_t^2 \mid \distrib_t} + \frac{\log(2/\delta)}{\lambda}. 
\end{align*}
Plugging in the form of \(Z_t\) and using \pref{eq:freedman1}, we get the desired statement. 
\end{proof}

Next we give a formal proof of \pref{lem:sq_loss_generalization}, which gives a generalization bound for least squares regression when the samples are adapted to an increasing filtration (and are not necessarily i.i.d.).  The proof follows similarly to \citet[Lemma A.11]{agarwal2019reinforcement}. 

\begin{lemma}[\pref{lem:sq_loss_generalization} restated: Least squares generalization bound]
\label{lem:sq_loss_generalization_restate} 
Let \(R > 0\), \(\delta \in (0, 1)\), we consider a sequential function estimation setting, with an instance space $\Xcal$ and target space $\Ycal$. Let \(\cH: \cX \mapsto [-R, R]\) be a class of real valued functions. Let \(\cD = \crl*{(x_1, y_1), \dots, (x_T, y_T)}\) be a dataset of \(T\) points where $x_t \sim \distrib_t = \distrib_t(x_{1:t-1},y_{1:t-1})$, and \(y_t\) is sampled via the conditional probability $p(\cdot \mid x_t)$:
\begin{align*}
y_t \sim  p(\cdot \mid x_t) \ldef{} h^*(x_t) + \varepsilon_t, 
\end{align*} where the function \(h^*\) satisfies approximate realizability i.e. $$\inf_{h \in \cH}  \frac{1}{T} \sum_{t=1}^T  \En_{x \sim \distrib_t} \brk*{\prn*{h^*(x) - h(x)}^2} \leq \gamma,$$ and \(\crl*{\epsilon_i}_{i=1}^n\)  are independent  random variables such that $\EE[y_t \mid x_t] = h^{\ast}(x_t)$. Additionally, suppose that \(\max_t \abs{y_t} \leq R\) and \(\max_{x} \abs*{h^*(x)} \leq R\). Then the least square solution \(\wh h \leftarrow \argmin_{h \in \cH} \sum_{t=1}^T \prn*{h(x_t) - y_t}^2\) satisfies with probability at least \(1 - \delta\), 
\begin{align*}
 \sum_{t=1}^T \En_{x \sim \distrib_t} \brk*{\prn{\wh h(x) - h^*(x)}^2} &\leq 3 \gamma T + 256 R^2 \log(2 \abs{\cH}/\delta). 
\end{align*}
\end{lemma} 
\begin{proof}  Consider any fixed function \(h \in \cH\) and define the random variable 
\begin{align*}
Z_t^h \ldef{} \prn*{h(x_t) - y_t}^2 - \prn*{h^*(x_t) - y_t}^2. 
\end{align*} 
Define the notation \(\En \brk*{\cdot \mid \distrib_t}\) to denote \(\En_{x_t \sim \distrib_t} \brk*{\cdot}\), and note that 
\begin{align*}
\En \brk*{Z^h_t \mid \distrib_t} &= \En_{x_t \sim \distrib_t} \brk*{\prn*{h(x_t) - h^*(x_t)}\prn*{h(x_t) + h^*(x_t) - 2 y_i}} = \En_{x_t \sim \distrib_t} \brk*{\prn*{h(x_t) - h^*(x_t)}^2}, \numberthis \label{eq:FQI_1}
\end{align*} where the last line holds because \(\En \brk*{y_t \mid x_t} = h^*(x_t)\). Furthermore, we also have that 
\begin{align*}
\En \brk*{(Z_t^h)^2 \mid \distrib_t} &= \En_{x_t \sim \distrib_t} \brk*{\prn*{h(x_t) - h^*(x_t)}^2 \prn*{h(x_t) + h^*(x_t) - 2 y_t}^2} \\ 
&\leq 16 R^2 \En_{x_t \sim \distrib_t} \brk*{\prn*{h(x_t) - h^*(x_t)}^2}. \numberthis \label{eq:FQI_2}
\end{align*}

Now we can note that the sequence of random variables \(\crl*{Z^h_1, \dots, Z^h_T}\) satisfies the condition in  \pref{lem:freedman_form} with\(\abs*{Z_t^h} \leq 4R^2\). Thus we get that for any \(\lambda \in [0, 1/8R^2]\) and \(\delta > 0\),  with probability at least \(1 - \delta\), 
\begin{align*}
\abs*{\sum_{t=1}^T Z^h_t -\En\brk*{Z^h_t\mid \distrib_t}} &\leq \lambda \sum_{t=1}^T \prn*{8  R^2 \abs*{\En \brk*{Z^h_t\mid \distrib_t} }  + \En\brk*{\prn*{Z^h_t}^2\mid \distrib_t}} + \frac{\log(2/\delta)}{\lambda}  \\
&\leq 32 \lambda R^2 \sum_{t=1}^T \En_{x_t \sim \distrib_t} \brk*{\prn*{h(x_t) - h^*(x_t)}^2} + \frac{\log(2/\delta)}{\lambda},  
\end{align*} where the last inequality uses \pref{eq:FQI_1} and \pref{eq:FQI_2}. Setting \(\lambda = 1/64 R^2\) in the above, and taking a union bound over \(h\), we get that for any \(h \in \cH\) and \(\delta > 0\), with probability at least \(1 - \delta\), 
\begin{align*}
\abs*{\sum_{t=1}^T Z^h_t -\En\brk*{Z^h_t \mid \distrib_t}} 
\leq \frac{1}{2} \sum_{t=1}^T \En_{x_t \sim \distrib_t} \brk*{\prn*{h(x_t) - h^*(x_t)}^2} + 64 R^2\log(2 \abs{\cH}/\delta). 
\end{align*} Rearranging the terms and using \pref{eq:FQI_1} in the above implies that, 
\begin{align*}
\sum_{t=1}^T Z_t^h  \leq \frac{3}{2}  \sum_{t=1}^T \En_{x_t \sim \distrib_t} \brk*{\prn*{h(x_t) - h^*(x_t)}^2}  + 64 R^2\log(2 \abs{\cH}/\delta) 
\intertext{and}
 \sum_{t=1}^T \En_{x_t \sim \distrib_t} \brk*{\prn*{h(x_t) - h^*(x_t)}^2} \leq 2 \sum_{t=1}^T Z_t^h  + 128 R^2 \log(2 \abs{\cH}/\delta).  \numberthis \label{eq:FQI_3}
\end{align*}
For the rest of the proof, we condition on the event that \pref{eq:FQI_3} holds for all \(h \in \cH\).  

Define the function \(\wt h \ldef{} \argmin_{h \in \cH} \sum_{t=1}^T \En_{x_t \sim \distrib_t} \brk*{(h(x_t) - h^*(x_t))^2}\). Using \pref{eq:FQI_3}, we get that 
\begin{align*}
\sum_{t=1}^T Z_t^{\wt h}  &\leq \frac{3}{2}  \sum_{t=1}^T \En_{x_t \sim \distrib_t} \brk*{\prn{\wt h(x_t) - h^*(x_t)}^2}  + 64 R^2\log(2 \abs{\cH}/\delta)  \\
&\leq \frac{3}{2} \gamma T + 64 R^2 \log(2 \abs{\cH}/\delta), 
\end{align*} where the last inequality follows from the approximate realizability assumption. Let \(\wh h\) denote the least squares solution on dataset \(\crl*{(x_t, y_t)}_{t \leq T}\). By definition, we have that
\begin{align*}
\sum_{t=1}^T Z_t^{\wh h} &= \prn{\wh h(x_t) - y_t}^2 - \prn{h^*(x_t) - y_t}^2 \leq \prn{\wt h(x_t) - y_t}^2 - \prn{h^*(x_t) - y_t}^2 = \sum_{t=1}^T Z_t^{\wt h}. 
\end{align*}
Combining the above two relations, we get that 
\begin{align*}
\sum_{t=1}^T Z_t^{\wh h} &\leq \frac{3}{2} \gamma T + 64 R^2 \log(2  \abs{\cH}/\delta).  \numberthis \label{eq:FQI_4}
\end{align*}

Finally, using \pref{eq:FQI_3} for the function \(\wh h\), we get that 
\begin{align*}
 \sum_{t=1}^T \En_{x_t \sim \distrib_t} \brk*{\prn{\wh h(x_t) - h^*(x_t)}^2} &\leq 2 \sum_{t=1}^T Z_t^{\wh h}  + 128 R^2 \log(2 \abs{\cH}/\delta) \\
 &\leq 3 \gamma T + 256 R^2 \log(2 \abs{\cH}/\delta), 
\end{align*}
where the last inequality uses the relation \pref{eq:FQI_4}. 
\end{proof}

\newcommand{\BEdim}{\mathrm{dim}_{\mathrm{DE}}}
\newcommand{\Belldim}{\mathrm{dim}_{\mathrm{BE}}}

\section{Low Bellman Eluder Dimension problems}  \label{app:Bellman_eluder} 

In this section, we consider problems with low Bellman Eluder dimensions \citet{jin2021bellman}.    This complexity measure is a distributional version of the Eluder dimension applied to the class of Bellman residuals w.r.t.~$\cF$. We show that our algorithm \algname{} gives a similar performance guarantee for problems with small Bellman Eluder dimensions. This demonstrates that \algname{} applies to any general model-free RL frameworks known in the RL literature so far.
 
We first introduce the key definitions:

\begin{definition}[{$\varepsilon$-independence between distributions \citep{jin2021bellman}}]
Let $\cG$ be a class of functions defined on a space $\cX$, and
$\nu, \mu_1,\dots,\mu_n$ be probability measures over $\cX$. We say $\nu$ is $\varepsilon$-independent of $\{\mu_1, \mu_2, \dots , \mu_n\}$ with respect to $\cG$ if there exists $g \in \cG$ such that $\sqrt{\sum_{i=1}^n(\bbE_{\mu_i}[g])^2}\leq \varepsilon$, but $|\bbE_\nu[g]| > \varepsilon$. 
\end{definition}

\begin{definition}[Distributional Eluder (DE) dimension]
Let $\cG$ be a function class defined on $\cX$ , and
$\cP$ be a family of probability measures over $\cX$ . The distributional Eluder dimension $\dim_{\operatorname{DE}}(\cF, \cP, \varepsilon)$
is the length of the longest sequence $\{\rho_1, \dots , \rho_n\} \subset \cP$ such that there exists $\varepsilon'\geq \varepsilon$ where $\rho_i$ is $\varepsilon'$-independent of $\{\rho_1, \dots, \rho_{i-1}\}$ for all $i \in [n]$.
\end{definition}

\begin{definition}[Bellman Eluder (BE) dimension \citep{jin2021bellman}]
Given a value function class \(\cF\), let \(\cG_h \ldef{} \prn*{f_h - \dynamicsopt f_{h+1} \mid f \in \cF_h, f_{h+1} \in \cF_{h+1}}\) be the
set of Bellman residuals induced by $\cF$ at step $h$, and $\cP = \{\cP_h\}_{h=1}^H$ be a collection of $H$ probability
measure families over $\cX \times \cA$. The $\epsilon$-Bellman Eluder dimension of $\cF$ with respect to $\cP$ is defined as
\begin{align*}
    \dim_{\operatorname{BE}}(\cF, \cP,\varepsilon) := \max_{h\in[H]}\dim_{\operatorname{DE}}(\cG_h, \cP_h,\epsilon)\,.
\end{align*}
\end{definition}

We also note the following lemma that controls the rate at which Bellman error accumulates. 
\begin{lemma}[Lemma 41, \citep{jin2021bellman}]  Given a function class \(\cG\) defined on a space \(\cX\) with \(\sup_{g \in \cG, x \in \cX}\abs*{g(x)} \leq C\), and a set of probability measures \(\cP\) over \(\cX\). Suppose that the sequence \(\crl*{g_k}_{k=1}^K \subset \cG\)  and \(\crl*{\mu_k}_{k=1}^K \subset \cP\) satisfy that \(\sum_{t=1}^{k-1} \prn*{\En_{\mu_t} \brk*{g_k}}^2 \leq \beta\) for all \(k \in [K]\). Then, for all \(k \in [K]\) and \(\gamma > 0\), 
\begin{align*} 
\sum_{t=1}^k \abs*{\En_{\mu_t} \brk*{g_t}} \leq O\prn*{\sqrt{\BEdim(\cG, \cP, \gamma) \beta k} + \min\crl*{k, \BEdim(\cG, \cP, \gamma) C} + k \gamma}. 
\end{align*} 
\label{lem:be_dim}
\end{lemma}

We next state our main theorem whose proof is similar to that of \pref{thm:main_regret}.

\begin{theorem}[Cumulative suboptimality]
\label{thm:main_regret_BE_dim} 
Fix $\delta\in (0,1)$, \(\moff = HT / d\) and \(\mon = H^2\), and suppose that the 
underlying MDP admits Bellman eluder dimention $d$, and the function class \(\cF\) satisfies~\pref{ass:Bellman_completeness}. Then with probability at least \(1 - \delta\),  \pref{alg:fqi} obtains the following bound on cumulative subpotimality w.r.t.~any comparator policy \(\pi^e\), 
\begin{align*}
 \sum_{t = 1}^T V^{\pi^e} - V^{\pi^{t}} = \widetilde{O}\prn*{ \Vmax \max\crl*{C_{\pi^e}, 1} \sqrt{d T \cdot \log\prn*{ H \abs{\cF}/\delta}}}, 
\end{align*}
where \(\pi^t = \pi^{f^t}\) is the greedy policy w.r.t. \(f^t\) at round \(t\) and \(d = \Belldim(\cF, \cP_{\cF}, 1/\sqrt{T})\). Here $\cP_{\Fcal}$ is the class of occupancy measures that can be be induced by greedy policies w.r.t. value functions in $\Fcal$.
\end{theorem}
\begin{proof} Repeating the analysis till \pref{eq:checkpoint_1} in the proof of \pref{thm:main_regret}, we get that 
\begin{align*}
  \sum_{t = 1}^T V^{\pi^e} - V^{\pi^{t}} &\leq T C_{\pi^e} \cdot  \sqrt{H \cdot \Delta_{\mathrm{off}}} + \sum_{t=1}^T \sum_{h = 0}^{H - 1} \abs*{ \EE_{s,a \sim d_h^{\pi^{f^t}}}\brk*{f^t_h(s,a) - \Tcal_{h}f^t_{h+1}(s,a)}} 
\end{align*}

Using the bound in \pref{lem:fqi_guarantee} and \pref{lem:be_dim} in the above, we get that 
\begin{align*}
  \sum_{t = 1}^T V^{\pi^e} - V^{\pi^{t}} &\lesssim T C_{\pi^e} \cdot  \sqrt{H \cdot \Delta_{\mathrm{off}}} +  \sum_{h = 0}^{H - 1} \sqrt{\BEdim(\cG_h, \cP_{\cF; h}, \gamma)  \Delta_{\mathrm{on}} T } \\
  &\qquad \qquad \qquad \qquad + \min \crl*{T, \BEdim(\cG_h, \cP_{\cF; h}, \gamma) C} + T \gamma.
\end{align*} where \(\cG_h \ldef{} \prn*{f_h - \dynamicsopt f_{h+1} \mid f \in \cF_h, f_{h+1} \in \cF_{h+1}}\) denotes the
set of Bellman residuals induced by $\cF$ at step $h$, and $\cP = \{\cP_{\cF; h}\}_{h=1}^H$ is the collection of occupancy measures at step \(h\) induced by greedy policies w.r.t.~value functions in \(\cF\). We set \(\gamma = 1/\sqrt{T}\) and define \(d = \Belldim(\cF, \cP, \gamma) = \max_{h} \BEdim(\cG_h, \cP_{\cF; h}, \gamma)\). Ignoring the lower order terms, we get that 
\begin{align*}
  \sum_{t = 1}^T V^{\pi^e} - V^{\pi^{t}} &\lesssim T C_{\pi^e} \cdot  \sqrt{H \cdot \Delta_{\mathrm{off}}} + H \sqrt{d  \Delta_{\mathrm{on}} T } \\
    &\lesssim T C_{\pi^e} \Vmax \cdot  \sqrt{H \cdot \frac{\log(H T \abs{\cF}/\delta)}{\moff}} + H \Vmax \sqrt{d T \cdot \frac{\log( H T \abs{\cF}/\delta)}{\mon}}, 
\end{align*} where \(\lesssim\) hides lower order terms, multiplying constants and log factors. Setting \(\moff = HT/d\) and \(\mon = H^2\), we get that 
\begin{align*}
  \sum_{t = 1}^T V^{\pi^e} - V^{\pi^{t}} &= \widetilde{O} \prn*{ C_{\pi^e}
  \Vmax \sqrt{d T \log( H T \abs{\cF}/\delta)}}. 
\end{align*}
\end{proof} 
\section{{Comparison with previous works}}\label{app:comp}

As mentioned in the main text, many previous empirical works consider combining offline expert demonstrations with online interaction \citep{rajeswaran2017learning,hester2018deep,nair2018overcoming,nair2020accelerating, vecerik2017leveraging, lee2022offline, jia2022improving, niu2022trust}. Thus the idea of performing RL algorithm on both offline data (expert demonstrations) and online data is also explored in some of the previous works, for example, \citet{vecerik2017leveraging} runs DDPG on both the online and expert data, and \citet{hester2018deep} uses DQN on both data but with an additional supervised loss. Since we already compared with \citet{hester2018deep} in the experiment, here we focus on our discussion with \citet{vecerik2017leveraging}. 

We first emphasize that \citet{vecerik2017leveraging} only focuses on expert demonstrations and their experiments entirely rely on using expert demonstrations, while we focus on more general offline dataset that is not necessarily coming from experts. Said though, the DDPG-based algorithm from \citet{vecerik2017leveraging} potentially can be used when offline data is not from experts. Although the algorithm from \citet{vecerik2017leveraging} and Hy-Q share the same high-level intuition that one should perform RL on both the datasets, there are still a few differences : (1) Hy-Q uses Q-learning instead of deterministic policy gradients; note that deterministic policy gradient methods cannot be directly applied to discrete action setting; (2) Hy-Q does not require n-step TD style update, since in off-policy case, without proper importance weighting, n-step TD could incur strong bias. While proper tuning on n could balance bias and variance, one does not need to tune such n-step at all in Hy-Q; (3) The idea of keeping a non-zero ratio to sample offline dataset is also proposed in \citet{vecerik2017leveraging}. Our buffer ratio is derived from our theory analysis but meanwhile proves the advantage of the similar heuristic applied in \citet{vecerik2017leveraging}.
(4) In their experiment, \citet{vecerik2017leveraging} only considers expert demonstrations. In our experiment, we considered offline datasets with different amounts of transitions from very low-quality policies and showed Hy-Q is robust to low-quality transitions in offline data. 
Note that some of the differences may seem minor on the implementation level, but they may be important to the theory.

Regarding the experiments, our experimental evaluation adds the following insights over those in \citet{vecerik2017leveraging}: (i) hybrid methods can succeed without expert data, (ii) hybrid methods can succeed in hard exploration discrete-action tasks, (iii) the core algorithm ($Q$-learning vs DDPG) is not essential although some details may matter. Due to the similarity between the two methods, we believe some of these insights may also translate to \citet{vecerik2017leveraging} and we expect that the choice between Hy-Q and Hy-DDPG will be environment specific, as it is with the purely online versions of these methods. In some situations, $Q$-learning works does not immediately imply Deterministic policy gradient methods work, nor vice versa. Nevertheless, it is beyond the scope of this paper to rigorously verify this claim and we deem the study of Actor-critic algorithms in Hybrid RL setting an interesting future direction. 

\section{Experiment Details}
\label{app:exp}
\subsection{Combination Lock}
In this section we provide a detailed description of combination lock experiment. The combination lock environment has a horizon $H$ and 10 actions at each state. There are three latent states $z_{i,h}, i \in \{0,1,2\}$ for each timestep $h$, where $z_{i,h}, i \in \{0,1\}$ are good states and $z_{2,h}$ is the bad state. For each good state, we randomly pick a good action $a_{i,h}$, such that in latent state $z_{i,h}, i \in \{0,1\}$, taking the good action $a_{i,h}$ will result in 0.5 probability of transiting to $z_{0, h+1}$ and 0.5 probability of transiting to $z_{1, h+1}$ while taking all other actions will result in a 1 probability of transiting to $z_{2, h+1}$. At $z_{2, h}$, all actions will result in a deterministic transition to $z_{2, h+1}$. For the reward, we give an optimal reward of 1 for landing $z_{i, H}, i \in \{0,1\}$. We also give an anti-shaped reward of 0.1 for all transitions from a good state to a bad state. All other transitions have a reward of 0. The initial distribution is a uniform distribution over $z_{0,0}$ and $z_{1,0}$. The observation space has dimension $2^{\left \lceil{\log(H+1)}\right \rceil }$, created by concatenating a one-hot representation of the latent state and a one-hot representation of the horizon (appending 0 if necessary). Random noise from $\mathcal{N}(0, 0.1)$ is added to each dimension, and finally the observation is multiplied by a Hadamard matrix. Note that in this environment, the agent needs to perform optimally for all $H$ timesteps to hit the final good state for an optimal reward of 1. Once the agent chooses a bad action, it will stay in the bad state until the end with at most 0.1 possible reward for the trajectory received while transitting from a good state to a bad state.

\subsection{Implementation Details of Combination Lock experiment}
We train $H$ separate Q-functions for all $H$ timesteps. Our function class consists of an encoder and a decoder. For the encoder, we feed the observation into one linear layer with 3 outputs, followed by a softmax layer to get a state-representation. This design of encoder is intended to learn a one-hot representation of the latent state. We take a Kronecker Product of the state-representation and the action, and feed the result to a linear layer with only one output, which will be our Q value. In order to stabilize the training, we warm-start the Q-function of timestep $h-1$ with the encoder from the Q-function at  timestep $h$ of the current iteration and the decoder from the Q-function at time step $h-1$ of the previous iteration, for each iteration of training.

One remark is that since combination lock belongs to Block MDPs, we require a V-type algorithm instead of the Q-type algorithm as shown in the main text. The only difference lies in the online sampling process: instead of sampling from $d^{\pi^t}_h$, for each $h$, we sample from $d_h^{\pi^t}\circ \textrm{Uniform}(\cA)$, i.e., we first rollin with respect to $\pi^t$ to timestep $h-1$, then take a random action, observe the transition and collect that tuple. See \pref{alg:vtype}.

For CQL, we implemented the variant of CQL-DQN and picked the peak in the learning curve to report in the main paper (so in other words, we pick the best performance of CQL using online samples). While CQL is supposed to be a pure offline RL baseline, we simply tune the hyperparameters of CQL using online samples. Thus our reported results can be regarded as an upper bound on the performance that CQL would get if it were trained in the pure offline setting.

\subsection{Implementation Details of Montezuma's Revenge experiment}
For Montezuma's Revenge, we follow the Atari game convention and use a discounted setting with discount factor being 0.99. 
In this section we provide the detailed algorithm for the discounted setting. The overall algorithm is described in \pref{alg:discounted}. For the function approximation, we use a class of convolutional neural networks (parameterized by class $\Theta$) as promoted by the original \textsc{Dqn} paper \citep{mnih2015human}. We include several standard empirical design choices that have been commonly used to stabilize the training: we use Prioritize Experience Replay \citep{schaul2015prioritized} for our buffer. We also add Double \textsc{Dqn} \citep{van2016deep} and Dueling \textsc{Dqn} \citep{wang2016dueling} during our Q-update. These are the standard heuristics that have been used in Q learning implementation commonly. We also observe that a decaying schedule on the offline sample ratio $\beta$ and the exploration rate $\epsilon$ also helps provide better performance. Note that an annealing $\beta$ does not contradict to our comment in \pref{sec:hyq} on catastrophic forgetting because $\beta$ always has a lower bound and indeed never decreases to zero when the  online learning process proceeds. In addition, we also perform value function update very $n_{\textrm{value}}$ inside each episode, instead of per episode update since this has been the popular design choice and leads to better efficiency in practice.

\begin{algorithm}[t] 
\caption{Discounted Hy-Q}
\begin{algorithmic}[1] 
\REQUIRE Value function class: $\Fcal$ (induced by $\Theta$), \#iterations: $T$, Offline dataset \(\cD^{\nu}\) of size \(m_\mathrm{off}\), discounted factor $\gamma$, value function update frequency $n_{\textrm{value}}$, target update frequency \(n_\mathrm{target}\), learning rate $\alpha$, offline sample ratio $\beta$, exploration rate $\epsilon$, action space $\Acal$. 
\\
\STATE Randomly initialize value function \(f^{\theta}\).
\STATE Initialize target value function \(\tilde f = f^{\theta}\).
\STATE Initialize online buffer $\Dcal = \emptyset$.
\STATE Sample initial state $s \sim d_0$
\FOR{$t = 1, \dots, T$}  
\STATE Let $\pi$ be the $\epsilon$-greedy policy w.r.t. \(f^{\theta}\) i.e., $\pi(s) = \argmax_a f^{\theta}(s,a)$ with probability $1-\epsilon$ and  $\pi(s) = \Ucal(\Acal)$ with probability $\epsilon$.
\\
 \vspace{1mm} 
\algcomment{Online collection} 
 \vspace{1mm} 
\\ 
\STATE Interact with the environment for one step:
$$a = \pi(s), s' \sim P(s,a), r \sim R(s,a).$$
\STATE Update online buffer: $\Dcal = \Dcal \cup \{s,a,r,s'\}$.
\\
 \vspace{1mm} 
 \algcomment{Discounted minibatch FQI using both online and offline data}
 \vspace{1mm}
 \IF{$t \mod n_{\textrm{value}} = 0$}
 \STATE With probability $1-\beta$: Sample a minibatch $D$ with size $n_\textrm{minibatch}$ from online buffer $\Dcal$.\\ Otherwise: Sample a minibatch $D$ with size $n_\textrm{minibatch}$ from offline buffer $\Dcal^{\nu}$.
\STATE Perform one-step gradient descent on $D$: 
\begin{align*}
    \theta = \theta - \alpha \nabla_{\theta}  \hat \EE_{D} \left(f^{\theta}(s, a) - r_i - \gamma \max_{a'} \tilde f (s',a')\right)^2.
\end{align*}
\ENDIF
\\
 \vspace{1mm} 
 \algcomment{Delayed update of target function every $n_{\textrm{target}}$ updates}
 \vspace{1mm}
\IF{$t \mod n_{\textrm{target}} = 0$}
\STATE Set target function to the current value function: $\tilde f = f^{\theta}$.
\ENDIF
\STATE Update $s \leftarrow s'$.
\ENDFOR
\end{algorithmic}\label{alg:discounted} 
\end{algorithm}

\subsection{Baseline implementation}
\subsubsection{Combination Lock}
We use the open-sourced implementation \url{https://github.com/BY571/CQL/tree/main/CQL-DQN} for \cql. For \briee, we use the official code released by the authors: \url{https://github.com/yudasong/briee}, where we rely on the code there for the combination lock environment.

\subsubsection{Montezuma's Revenge}
We use the open-sourced implementation  \\\url{https://github.com/jcwleo/random-network-distillation-pytorch} for \rnd. For \cql, we use \url{https://github.com/takuseno/d3rlpy} for their implementation of \cql for atari. We use \url{https://github.com/felix-kerkhoff/DQfD} for \dqfd. For all baselines, we keep the hyperparameters used in these public repositories. For \cql and \dqfd, we provide the offline datasets as described in the main text instead of using the offline dataset provided in the public repositories.\footnote{We note that \cql also fails completely with the original offline dataset (with 1 million samples) provided in the public repository.} All baselines are tested in the same stochastic environment setup as in \citet{burda2018exploration}.

\subsection{Hardware Infrastructure}
We run our experiments on a cluster of computes with Nvidia RTX 3090 GPUs and various CPUs which do not incur any randomness to the results. 

\subsection{Hyperparameters}
\subsubsection{Combination Lock}
We provide the hyperparameters of Hy-Q in Table.~\pref{table:hybrid_combination_lock}. In addition, we provide the hyperparameters we tried for \cql baseline in Table.~\ref{table:cql}.
\begin{table}[h] 
\caption{Hyperparameters for Hy-Q in combination lock}
\centering
\begin{tabular}{ccc} 
\toprule
                                                &\; Value Considered          &\; Final Value  \\ 
\hline
Learning rate                                   &\; \{1e-2, 2e-2, 1e-3\}                  &\; 2e-2         \\
Buffer size                                     &\; \{1e8\}                  &\; 1e8         \\
Optimizer                                 &\; \{Adam, SGD\}                  &\; Adam       \\
Number of updates per iteration                 &\; \{30, 300, 500\}                   &\; 500         \\
Batch size                                      &\; \{512\}                    &\; 512           \\
\toprule
\end{tabular}\label{table:hybrid_combination_lock}\end{table}

\begin{table}[h] 
\caption{Hyperparameters for CQL(DQN) in combination lock}
\centering
\begin{tabular}{ccc} 
\toprule
                                                &\; Value Considered          &\; Final Value  \\ 
\hline
Learning rate                                   &\; \{1e-3\}                  &\; 1e-3         \\
Optimizer                                 &\; \{Adam\}                  &\; Adam       \\
Buffer size                                     &\; \{1e8\}                  &\; 1e8         \\
Batch size                                      &\; \{512\}                    &\; 512           \\
Discount Factor              &\; \{0.99\}                   &\; 0.99        \\
Moving Average Factor $\tau$              &\; \{0.01, 0.1, 1\}                   &\; 0.01        \\
Weight on CQL loss $\alpha$              &\; \{0, 0.1, 0.01\}                   &\; 0.1        \\
\toprule
\end{tabular}
\label{table:cql}
\end{table}

\subsubsection{Montezuma's Revenge}
We provide the hyperparameter of Hy-Q in Table.~\ref{app:table:hyperparam:mr}. We reuse many hyperparameter choices from \textsc{Dqfd}. Note that $[a,b]$ denotes a decreasing/increasing schedule from $a$ to $b$.
\begin{table}[h] 
\caption{Hyperparameter of Discounted Hy-Q in Montezuma's Revenge.}
\centering
\begin{tabular}{ccc} 
\toprule
                                                &\; Value Considered            &\; Final Value  \\ 
\hline
Learning rate                                   &\; \{6.25e-5, [1e-4,1e-5]\}    &\; [1e-4,1e-5]         \\
Offline Schedule $\beta$                        &\; \{0.5,0.2,[0.2,0.01]\}      &\; [0.2,0.01]         \\
Exploration $\epsilon$ rate                     &\; \{[0.25,0.001]\}            &\; [0.25,0.001]          \\
Minibatch size $n_\textrm{minibatch}$           &\; \{32\}                      &\; 32           \\
Weight decay (regularization) coefficient       &\; \{1e-5\}                    &\; 1e-5           \\
Gradient Clipping                               &\; \{10,20\}                   &\; 10          \\
Discount factor $\gamma$                        &\; \{0.99\}                    &\; 0.99          \\
Value function update frequency $n_\textrm{value}$ &\; \{4\}                   &\; 4          \\
Target function update frequency $n_\textrm{target}$    &\; \{1000,2000,5000,10000\}           &\; 10000          \\
Buffer size                                     &\; \{$2^{20}$\}                &\;$2^{20}$         \\
PER Importance Sampling ratio                   &\; \{[0.6,1]\}                 &\;[0.6,1]         \\
Online PER $\epsilon$                           &\; \{0.001\}                   &\;0.001         \\
Offline PER $\epsilon$                          &\; \{0.0001\}                  &\;0.0001         \\
Online PER Priority Coefficient                 &\; \{0.4\}                     &\;0.4         \\
Offline PER Priority Coefficient                &\; \{1\}                       &\;1         \\
\toprule
\end{tabular}
\label{app:table:hyperparam:mr}
\end{table}

\end{document}